\documentclass[11pt,table]{article}
\pdfoutput=1

\usepackage[utf8]{inputenc} 
\usepackage[T1]{fontenc}    
\usepackage[hyphens]{url}            
\usepackage[in]{fullpage}
\usepackage[hyphenbreaks]{breakurl}
\usepackage{hyperref}       
\usepackage{booktabs}       
\usepackage{amsfonts}       
\usepackage{nicefrac}       
\usepackage{microtype}      
\usepackage{amsmath,amssymb,amsthm}
\usepackage{mathtools}
\usepackage{mathrsfs}
\usepackage{thmtools}
\usepackage{thm-restate}
\usepackage{comment}
\usepackage{parskip}
\usepackage{etoolbox}
\usepackage{enumitem}
\usepackage[textsize=scriptsize,textwidth=3cm]{todonotes}
\usepackage{subcaption}
\usepackage{tabularx}
\usepackage{xcolor} 
\usepackage[numbers,sort]{natbib}

\newtoggle{showtodos}
\toggletrue{showtodos}

\newtoggle{isnips}
\togglefalse{isnips}

\newcommand{\dist}{\ensuremath{\mathcal{D}}}
\newcommand{\fhat}{\ensuremath{\hat{f}}}
\newcommand{\ermloss}{\ensuremath{L}}
\newcommand{\Eop}{\ensuremath{\mathop{\mathbb{E}}}}
\newcommand{\Dtest}{\ensuremath{D_{\text{test}}}}
\newcommand{\D}[1]{\ensuremath{D_{\text{#1}}}}
\newcommand{\ind}{\ensuremath{\mathbb{I}}}
\newcommand{\accorig}{\ensuremath{\text{acc}_\text{orig}}}
\newcommand{\accnew}{\ensuremath{\text{acc}_\text{new}}}

\newcommand{\keyword}[1]{\texttt{#1}}
\newcommand{\model}[1]{\texttt{#1}}
\newcommand{\class}[1]{\texttt{#1}}
\newcommand{\airplane}{\class{airplane}}

\newcolumntype{L}[1]{>{\raggedright\arraybackslash}p{#1}}
\newcolumntype{C}[1]{>{\centering\arraybackslash}p{#1}}
\newcolumntype{R}[1]{>{\raggedleft\arraybackslash}p{#1}}

\DeclarePairedDelimiter{\abs}{\lvert}{\rvert}

\DeclarePairedDelimiter{\brackets}{[}{]}

\iftoggle{showtodos}{
\newcommand{\becca}[1]{\todo[color=green!40]{Becca: #1}}
\newcommand{\ludwig}[1]{\todo[color=blue!40]{Ludwig: #1}}
}{
\newcommand{\becca}[1]{}
\newcommand{\ludwig}[1]{}
}

\title{Do CIFAR-10 Classifiers Generalize to CIFAR-10?}
\author{Benjamin Recht\\ UC Berkeley\and Rebecca Roelofs\\ UC Berkeley\and Ludwig Schmidt\\ MIT\and Vaishaal Shankar\\ UC Berkeley}

\begin{document}

\maketitle

\begin{abstract}
Machine learning is currently dominated by largely experimental work focused on improvements in a few key tasks.
However, the impressive accuracy numbers of the best performing models are questionable because the same test sets have been used to select these models for multiple years now.
To understand the danger of overfitting, we measure the accuracy of CIFAR-10 classifiers by creating a new test set of truly unseen images.
Although we ensure that the new test set is as close to the original data distribution as possible, we find a large drop in accuracy (4\% to 10\%) for a broad range of deep learning models.
Yet, more recent models with higher original accuracy show a \emph{smaller} drop and better overall performance, indicating that this drop is likely not due to overfitting based on adaptivity.
Instead, we view our results as evidence that current accuracy numbers are brittle and susceptible to even minute natural variations in the data distribution.
\end{abstract}

\section{Introduction}
\label{sec:intro}

Over the past five years, machine learning has become a decidedly experimental field.
Driven by a surge of research in deep learning, the majority of published papers has embraced a paradigm where the main justification for a new learning technique is its improved performance on a few key benchmarks.
At the same time, there are few explanations as to \emph{why} a proposed technique is a reliable improvement over prior work.
Instead, our sense of progress largely rests on a small number of standard benchmarks such as CIFAR-10, ImageNet, or MuJoCo.
This raises a crucial question:

\begin{center}
\emph{How reliable are our current measures of progress in machine learning?}
\end{center}

Properly evaluating progress in machine learning is subtle.
After all, the goal of a learning algorithm is to produce a model that generalizes well to \emph{unseen data}.
Since we usually do not have access to the ground truth data distribution, we instead evaluate a model's performance on a separate test set.
This is indeed a principled evaluation protocol, \emph{as long as we do not use the test set to select our models.}

Unfortunately, we typically have limited access to new data from the same distribution.
It is now commonly accepted to re-use the same test set multiple times throughout the algorithm and model design process.
Examples of this practice are abundant and include both tuning hyperparameters (number of layers, etc.) within a single publication, and building on other researchers' work across publications.
While there is a natural desire to compare new models to previous results, it is evident that the current research methodology undermines the key assumption that the classifiers are independent of the test set.
This mismatch presents a clear danger because the research community could easily be designing models that only work well on the specific test set but actually fail to generalize to new data \cite{BH15}.

\subsection{A Reproducibility Study on CIFAR-10}
To understand how reliable current progress in machine learning is, we design and conduct a new type of reproducibility study.
Its main goal is to measure how well contemporary classifiers generalize to new, truly unseen data from the same distribution.
We focus on the standard CIFAR-10 dataset since its transparent creation process makes it particularly well suited to this task. 
Moreover, CIFAR-10 has been the focus of intense research for almost 10 years now.
Due to the competitive nature of this process, it is an excellent test case for investigating whether adaptivity has led to overfitting.

Our study proceeds in three steps:
\iftoggle{isnips}{
\begin{enumerate}[leftmargin=*]
}{
\begin{enumerate}
}
\item First, we curate a new test set where we carefully match the \emph{sub}-class distribution of our new test set to the original CIFAR-10 dataset.
\item After collecting about 2000 new images, we evaluate the performance of 30 image classification models on our new test set.
The results show two overarching phenomena.
On the one hand, there is a significant drop in accuracy from the original test set to our new test set.
For instance, VGG and ResNet architectures \cite{vgg,resnet} drop from their well-established 93\% accuracy to about 85\% on our new test set.
On the other hand, we find the performance on the existing test set to be highly predictive of the performance on our new test set.
Even small incremental improvements on CIFAR-10 often transfer to truly held-out data.

\item Motivated by this discrepancy between the original and new accuracies, the third step investigates multiple hypotheses for explaining this gap.
A natural conjecture is that re-tuning standard hyperparameters recovers some of the observed gap, but we find only a small effect of about 0.6\% improvement.
While this and further experiments can explain some of the accuracy loss, a significant gap remains.
\end{enumerate}

Overall, our results paint an unexpected picture of progress in contemporary machine learning.
In spite of adapting to the CIFAR-10 test set for several years, there has been no stagnation.
The top model is still a recent Shake-Shake network with Cutout regularization \cite{shakeshake,cutout}.
Moreover, its advantage over a standard ResNet \emph{increased} from 4\% to 8\% on our new test set.
This shows that the current research methodology of ``attacking'' a test set for an extended period of time is surprisingly resilient to overfitting.

But our results also cast doubt on the robustness of current classifiers.
While our new dataset presents only a minute distributional shift, the classification accuracy of widely used models drops significantly.
For instance, the aforementioned accuracy loss of VGG and ResNet architectures corresponds to multiple years of progress on CIFAR-10 \cite{HSKSS12}.
Note that the distributional shift induced by our experiment is neither adversarial nor the result of a different data source.
So even in benign settings, distribution shift poses a serious challenge and questions to what extent current models truly generalize.

\section{Formal Setup}
\label{sec:formal}
Before we describe our specific experiment on CIFAR-10, we start with a formal description of our problem of interest.
We adopt the standard classification setup and posit the existence of a ``true'' underlying data distribution $\dist$ over labeled examples $(x, y)$.
The goal is to find a model $\fhat$ that minimizes the population loss
\begin{equation}
  \label{eq:pop_loss}
  \ermloss_\dist(\fhat) \; = \; \Eop_{(x, y) \sim \dist} \brackets*{\ind\brackets{\fhat(x) \neq y}} \; .
\end{equation}

Since we usually do not know the distribution $\dist$, we instead measure the performance of a trained classifier via a \emph{test set} $D_\text{test}$ drawn from the distribution $\dist$:
\begin{equation}
  \label{eq:emp_loss}
  \ermloss_{\Dtest}(\fhat) \; = \; \frac{1}{\abs{\Dtest}} \sum_{(x, y) \in \Dtest} \ind\brackets{\fhat(x) \neq y} \; .
\end{equation}
For a sufficiently large test set $\Dtest$, standard concentration results show that $\ermloss_{\Dtest}(\fhat)$ is a good approximation of $\ermloss_\dist(\fhat)$ as long as the classifier $\fhat$ does not depend on $\Dtest$.
This is arguably the core assumption underlying machine learning since it allows us to argue that our classifier $\fhat$ truly \emph{generalizes} (as opposed to say only memorizing the data).
So if we collect a new test set $\Dtest'$ from the same distribution $\dist$, we would expect that
the accuracies match up to confidence intervals given by the inherent sampling error:
\[
  \ermloss_{\dist}(\fhat) \; \approx \; \ermloss_{\Dtest}(\fhat) \; \approx \; \ermloss_{\Dtest'}(\fhat) \; .
\]
However, it is often hard to argue when a new test set is drawn from exactly the same distribution $\dist$ since we usually lack a precise definition of this distribution.
So to obtain truly i.i.d.\ test sets, we ideally would have collected a larger initial dataset that we then randomly split into $\D{train}$, $\Dtest$, and $\Dtest'$.
Unfortunately, we usually do not have such an exact setup to reproduce accuracy numbers on a new test set.
In this paper, we instead mimic the data generating distribution $\dist$ as closely as
possible by repeating the dataset creation process that originally derived $\D{train}$ and $\Dtest$
from a larger dataset. 
While this method does not necessarily generate a test set that is an i.i.d.\ draw from the original
data generating distribution, it is a close approximation. 

\section{Dataset Creation Methodology}

To investigate how well current image classifiers generalize to truly unseen data, we collect a new test set for the CIFAR-10 image classification dataset \cite{krizhevsky2009learning}.
There are multiple reasons for this choice:
\iftoggle{isnips}{
\begin{itemize}[leftmargin=*]
}{
\begin{itemize}
}
\item CIFAR-10 is currently one of the most widely used datasets in machine learning and serves as a test ground for many computer vision methods.
  A concrete measure of popularity is the fact that CIFAR-10 was the second most common dataset in
NIPS 2017 (after MNIST) \cite{hamner2018popular}.
\item The dataset creation process for CIFAR-10 is transparent and well documented \cite{krizhevsky2009learning}.
  Importantly, CIFAR-10 draws from the larger Tiny Images repository that has significantly more fine-grained labels \cite{tinyimages}.
This makes it possible to conduct an experiment where we minimize various forms of distribution shift in our new test set.
\item CIFAR-10 poses a difficult enough problem so that the dataset is still the subject of active research (e.g., see \cite{cutout,shakeshake,shakedrop,nas,RAHL18}).
Moreover, there is a wide range of classification models that achieve significantly different accuracy scores.
Since code for these models is published in a variety of open source repositories, they can be treated as truly independent of our new test set.
\end{itemize}

\subsection{Background}

Before we describe how we created our new test set, we briefly review relevant background on CIFAR-10 and Tiny Images.

\paragraph{Tiny Images.} 
The dataset contains 80 million RGB color images with resolution 32 $\times$ 32 pixels.
The images are organized by roughly 75,000 \emph{keywords} that correspond to the non-abstract nouns from the WordNet database.
Each keyword was entered into multiple Internet search engines to collect roughly 1,000 to 2,500 images per keyword.
It is important to note that Tiny Images is a fairly noisy dataset.
Many of the images filed under a certain keyword do not clearly (or not at all) correspond to the respective keyword.

\paragraph{CIFAR-10.} The goal for the CIFAR-10 dataset was to create a cleanly labeled subset of Tiny Images.
To this end, the researchers assembled a dataset consisting of ten classes with 6,000 images per class.  
These classes are \keyword{airplane}, \keyword{automobile}, \keyword{bird}, \keyword{cat}, \keyword{deer}, \keyword{dog}, \keyword{frog}, \keyword{horse}, \keyword{ship},  and \keyword{truck}.
The standard train / test split is class-balanced and contains 50,000 training images and 10,000 test images.

The CIFAR-10 creation process is well-documented \cite{krizhevsky2009learning}.
First, the researchers assembled a set of relevant keywords for each class by using the hyponym relations in WordNet \cite{wordnet}.
Since directly using the corresponding images from Tiny Images would not give a high quality dataset, the researchers paid student annotators to label the images from Tiny Images.
The labeler instructions can be found in Appendix C of \cite{krizhevsky2009learning} and include a set of specific guidelines (e.g., an image should not contain two object of the corresponding class).
The researchers then verified the labels of the images selected by the annotators and removed near-duplicates from the dataset via an $\ell_2$ nearest neighbor search.

\subsection{Building the New Test Set}
\label{sec:building_new_test_set}
Our overall goal was to create a new test set that is as close as possible to being drawn from the same distribution as the original CIFAR-10 dataset.
One crucial aspect here is that the CIFAR-10 dataset did not exhaust any of the Tiny Image keywords it is drawn from.
So by collecting new images from the same keywords as CIFAR-10, our new test set can match the sub-class distribution of the original dataset.

\paragraph{Understanding the Sub-Class Distribution.} As the first step, we determined the Tiny Image keyword for every image in the CIFAR-10 dataset.
A simple nearest-neighbor search sufficed since every image in CIFAR-10 had an exact duplicate ($\ell_2$-distance $0$) in Tiny Images.
Based on this information, we then assembled a list of the 25 most common keywords for each class.
We decided on 25 keywords per class since the 250 total keywords make up more than 95\% of CIFAR-10.
Moreover, we wanted to avoid accidentally creating a harder dataset with infrequent keywords that the classifiers had little incentive to learn based on the original CIFAR-10 dataset.

The keyword distribution can be found in Appendix \ref{apx:v4_keywords}.
Inspecting this list reveals the importance of matching the sub-class distribution.
For instance, the most common keyword in the \airplane{} class is \keyword{stealth\_bomber} and not an arguably more ordinary civilian type of airplane.
In addition, the third most common keyword for the \airplane{} class is \keyword{stealth\_fighter}.
Both types of planes are highly distinctive.
There are more examples where certain sub-classes are considerably different, e.g., images from the \keyword{fire\_truck} keyword have image statistics that are rather different from say pictures for \keyword{dump\_truck}.

\paragraph{Collecting New Images.} After determining the keywords, we collected corresponding images.
To simulate the student / researcher split in the original CIFAR-10 collection procedure, we introduced a similar split among two authors of this paper.
Author A took the role of the original student annotators and selected new suitable images for the 250 keywords.
In order to ensure a close match between the original and new images for each keyword, we built a user interface that allowed Author A to first look through existing CIFAR-10 images for a given keyword and then select new candidates from the remaining pictures in Tiny Images.
Author A followed the labeling guidelines in the original instruction sheet \cite{krizhevsky2009learning}.
The number of images Author A selected per keyword was so that our final dataset would contain between 2,000 and 4,000 images.
We decided on 2,000 images as a target number for two reasons:
\iftoggle{isnips}{
\begin{itemize}[leftmargin=*]
}{
\begin{itemize}
}
\item While the original CIFAR-10 test set contains 10,000 images, a test set of size 2,000 is already sufficient for a fairly small confidence interval.
In particular, a conservative confidence interval (Clopper-Pearson at confidence level 95\%) for accuracy 90\% has size about $\pm 1\%$ with $n =$ 2,000 (to be precise, $[88.6\%, \, 91.3\%]$).
Since we considered a potential discrepancy between original and new test accuracy only interesting if it was significantly larger than 1\%, we decided that a new test set of size 2,000 was large enough for our study.
\item As with very infrequent keywords, our goal was to avoid accidentally creating a harder test set.
Since some of the Tiny Image keywords have only a limited supply of remaining adequate images, we decided that a smaller target size for the new dataset would reduce bias to include images of more questionable difficulty.
\end{itemize}
After Author A had selected a set of about 9,000 candidate images, Author B adopted the role of the researchers in the original CIFAR-10 dataset creation process.
In particular, Author B reviewed all candidate images and removed images that were unclear to Author B or did not conform to the labeling instructions in their opinion (some of the criteria are subjective).
In the process, a small number of keywords did not have enough images remaining to reach the $n =$ 2,000 threshold.
Author B then notified Author A about the respective keywords and Author A selected a further set of images for these keywords.
In this process, there was only one keyword where Author A had to carefully go through all available images in Tiny Images.
This keyword was \keyword{alley\_cat} and comprises less than 0.3\% of the overall CIFAR-10 dataset.

\paragraph{Final Assembly.}
After collecting a sufficient number of high-quality images for each keyword, we sampled a random subset from our pruned candidate set.
The sampling procedure was such that the keyword-level distribution of our new dataset matches the
keyword-level distribution of CIFAR-10 (see Appendix \ref{apx:v4_keywords}).
In the final stage, we again proceeded similar to the original CIFAR-10 dataset creation process and used $\ell_2$-nearest neighbors to filter out near duplicates.
In particular, we removed near-duplicates within our new dataset and also images that had a near duplicate in the original CIFAR-10 dataset (train or test).
The latter aspect is particularly important since our reproducibility study is only interesting if we evaluate on truly unseen data.
Hence we manually reviewed the top-10 nearest neighbors for each image in our new test set.
After removing near-duplicates in our dataset, we re-sampled the respective keywords until this process converged to our final dataset.

We remark that we did not run any classifiers on our new dataset during the data collection phase of our study.
In order to ensure that the new data does not depend on the existing classifiers, it is important to strictly separate the data collection phase from the following evaluation phase.

\begin{figure}
\centering
\newlength{\imagedim}
\setlength{\imagedim}{1.2cm}
\newlength{\imagexspacing}
\setlength{\imagexspacing}{0.1cm}
\newlength{\imageyspacing}
\setlength{\imageyspacing}{0.1cm}
\begin{subfigure}[t]{0.49\textwidth}
\centering
\begin{tikzpicture}
\tikzstyle{img}=[inner sep=0pt,outer sep=0pt];
\node [img] (image0) {\includegraphics[width=\imagedim]{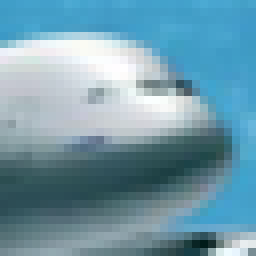}};
\node [img,anchor=west,at=(image0.east),xshift=\imagexspacing] (image1)
{\includegraphics[width=\imagedim]{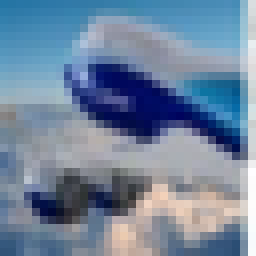}};
\node [img,anchor=west,at=(image1.east),xshift=\imagexspacing] (image2)
{\includegraphics[width=\imagedim]{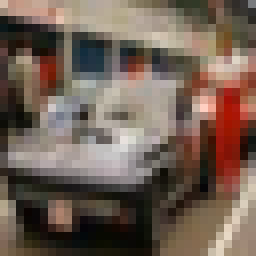}};
\node [img,anchor=west,at=(image2.east),xshift=\imagexspacing] (image3)
{\includegraphics[width=\imagedim]{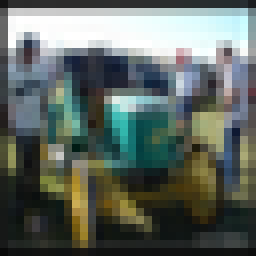}};
\node [img,anchor=west,at=(image3.east),xshift=\imagexspacing] (image4)
{\includegraphics[width=\imagedim]{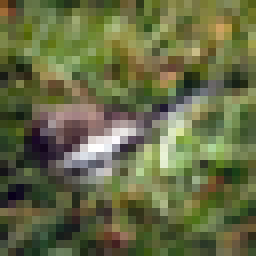}};
\node [img,anchor=north,at=(image0.south),yshift=-\imageyspacing] (image5)
{\includegraphics[width=\imagedim]{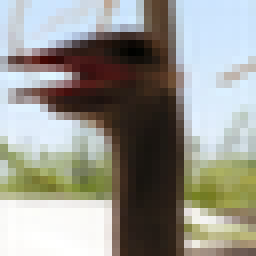}};
\node [img,anchor=west,at=(image5.east),xshift=\imagexspacing] (image6)
{\includegraphics[width=\imagedim]{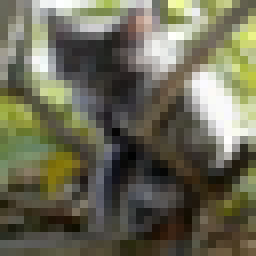}};
\node [img,anchor=west,at=(image6.east),xshift=\imagexspacing] (image7)
{\includegraphics[width=\imagedim]{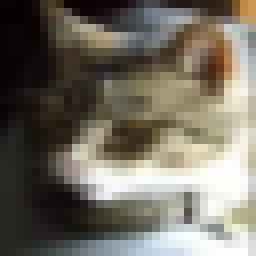}};
\node [img,anchor=west,at=(image7.east),xshift=\imagexspacing] (image8)
{\includegraphics[width=\imagedim]{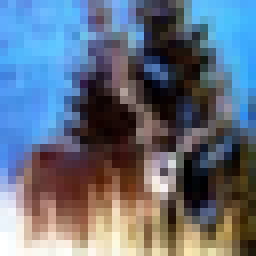}};
\node [img,anchor=west,at=(image8.east),xshift=\imagexspacing] (image9)
{\includegraphics[width=\imagedim]{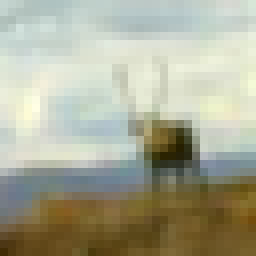}};
\node [img,anchor=north,at=(image5.south),yshift=-\imageyspacing] (image10)
{\includegraphics[width=\imagedim]{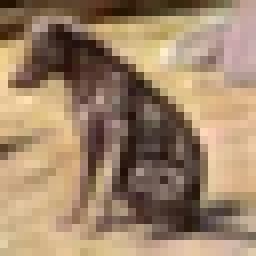}};
\node [img,anchor=west,at=(image10.east),xshift=\imagexspacing] (image11)
{\includegraphics[width=\imagedim]{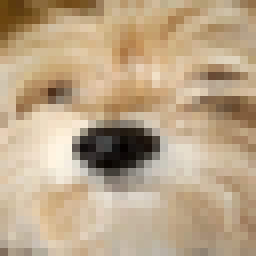}};
\node [img,anchor=west,at=(image11.east),xshift=\imagexspacing] (image12)
{\includegraphics[width=\imagedim]{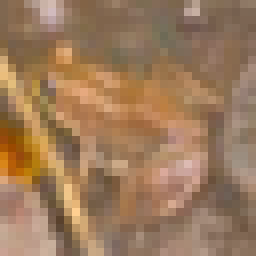}};
\node [img,anchor=west,at=(image12.east),xshift=\imagexspacing] (image13)
{\includegraphics[width=\imagedim]{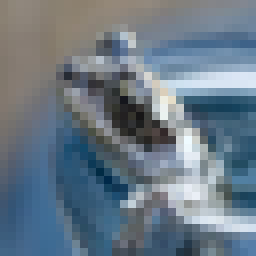}};
\node [img,anchor=west,at=(image13.east),xshift=\imagexspacing] (image14)
{\includegraphics[width=\imagedim]{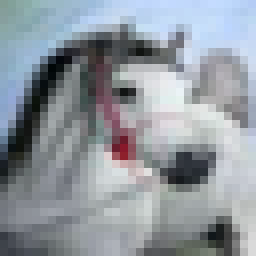}};
\node [img,anchor=north,at=(image10.south),yshift=-\imageyspacing] (image15)
{\includegraphics[width=\imagedim]{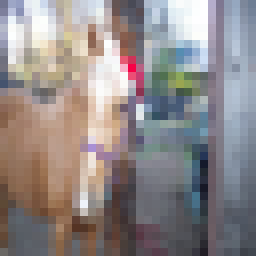}};
\node [img,anchor=west,at=(image15.east),xshift=\imagexspacing] (image16)
{\includegraphics[width=\imagedim]{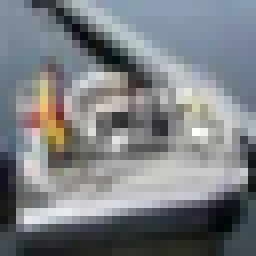}};
\node [img,anchor=west,at=(image16.east),xshift=\imagexspacing] (image17)
{\includegraphics[width=\imagedim]{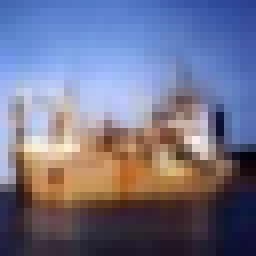}};
\node [img,anchor=west,at=(image17.east),xshift=\imagexspacing] (image18)
{\includegraphics[width=\imagedim]{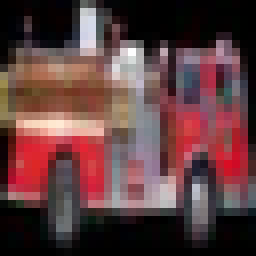}};
\node [img,anchor=west,at=(image18.east),xshift=\imagexspacing] (image19)
{\includegraphics[width=\imagedim]{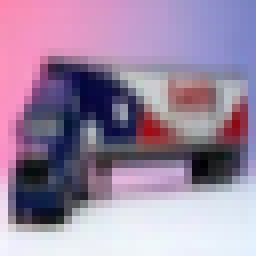}};
\end{tikzpicture}
  \caption{Test Set A}
  \label{fig:new_test}
\end{subfigure}
\begin{subfigure}[t]{0.49\textwidth}
\centering
\begin{tikzpicture}
\tikzstyle{img}=[inner sep=0pt,outer sep=0pt];
\node [img] (image0) {\includegraphics[width=\imagedim]{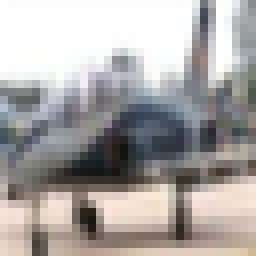}};
\node [img,anchor=west,at=(image0.east),xshift=\imagexspacing] (image1)
{\includegraphics[width=\imagedim]{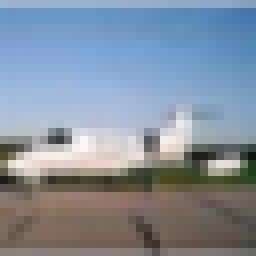}};
\node [img,anchor=west,at=(image1.east),xshift=\imagexspacing] (image2)
{\includegraphics[width=\imagedim]{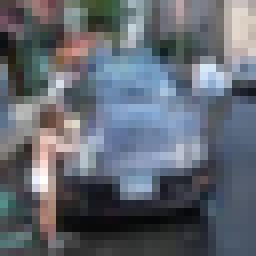}};
\node [img,anchor=west,at=(image2.east),xshift=\imagexspacing] (image3)
{\includegraphics[width=\imagedim]{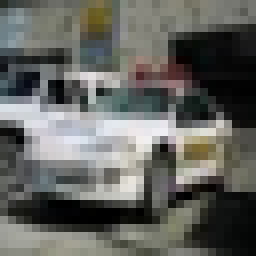}};
\node [img,anchor=west,at=(image3.east),xshift=\imagexspacing] (image4)
{\includegraphics[width=\imagedim]{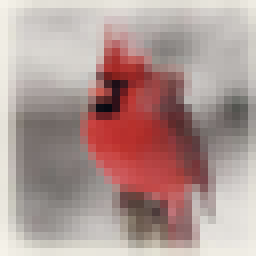}};
\node [img,anchor=north,at=(image0.south),yshift=-\imageyspacing] (image5)
{\includegraphics[width=\imagedim]{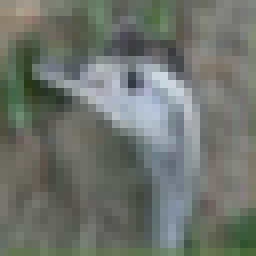}};
\node [img,anchor=west,at=(image5.east),xshift=\imagexspacing] (image6)
{\includegraphics[width=\imagedim]{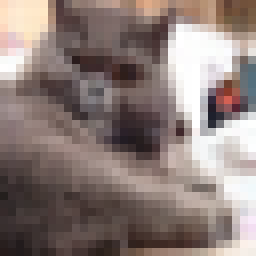}};
\node [img,anchor=west,at=(image6.east),xshift=\imagexspacing] (image7)
{\includegraphics[width=\imagedim]{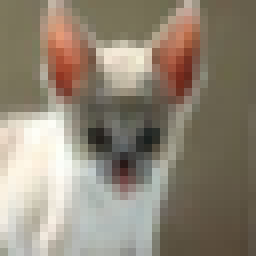}};
\node [img,anchor=west,at=(image7.east),xshift=\imagexspacing] (image8)
{\includegraphics[width=\imagedim]{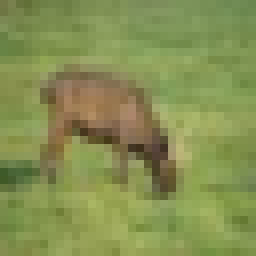}};
\node [img,anchor=west,at=(image8.east),xshift=\imagexspacing] (image9)
{\includegraphics[width=\imagedim]{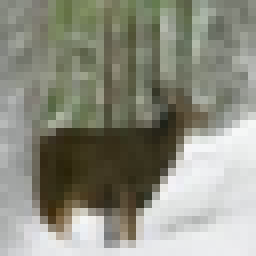}};
\node [img,anchor=north,at=(image5.south),yshift=-\imageyspacing] (image10)
{\includegraphics[width=\imagedim]{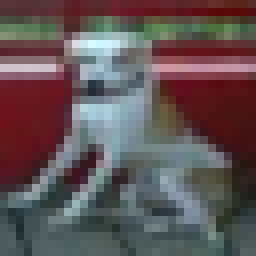}};
\node [img,anchor=west,at=(image10.east),xshift=\imagexspacing] (image11)
{\includegraphics[width=\imagedim]{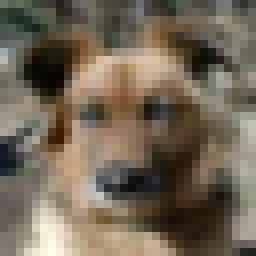}};
\node [img,anchor=west,at=(image11.east),xshift=\imagexspacing] (image12)
{\includegraphics[width=\imagedim]{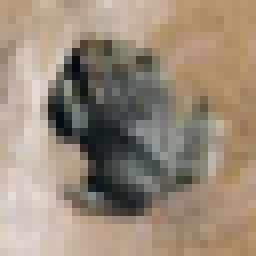}};
\node [img,anchor=west,at=(image12.east),xshift=\imagexspacing] (image13)
{\includegraphics[width=\imagedim]{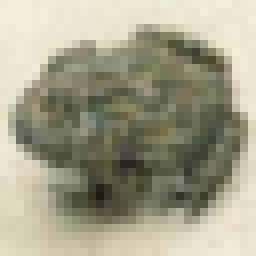}};
\node [img,anchor=west,at=(image13.east),xshift=\imagexspacing] (image14)
{\includegraphics[width=\imagedim]{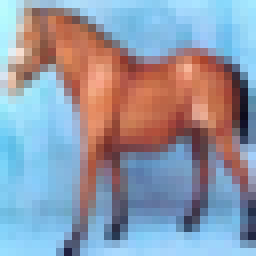}};
\node [img,anchor=north,at=(image10.south),yshift=-\imageyspacing] (image15)
{\includegraphics[width=\imagedim]{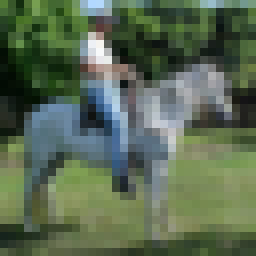}};
\node [img,anchor=west,at=(image15.east),xshift=\imagexspacing] (image16)
{\includegraphics[width=\imagedim]{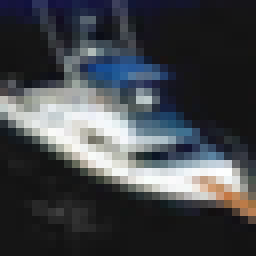}};
\node [img,anchor=west,at=(image16.east),xshift=\imagexspacing] (image17)
{\includegraphics[width=\imagedim]{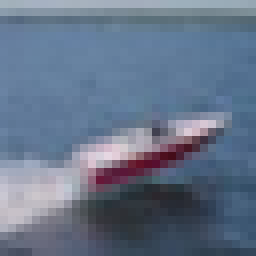}};
\node [img,anchor=west,at=(image17.east),xshift=\imagexspacing] (image18)
{\includegraphics[width=\imagedim]{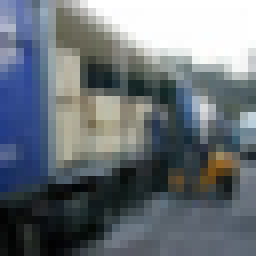}};
\node [img,anchor=west,at=(image18.east),xshift=\imagexspacing] (image19)
{\includegraphics[width=\imagedim]{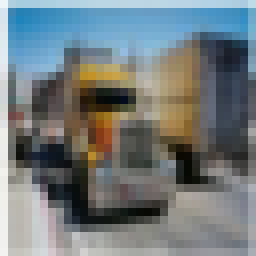}};
\end{tikzpicture}
  \caption{Test Set B}
  \label{fig:original_test}
\end{subfigure}
\caption{Class-balanced random draws from the new and original test sets.\protect \footnotemark}
\label{fig:testexamples}
\end{figure}
\footnotetext{Test Set A is the new test set and Test Set B is the original test set.}

\section{Model Performance Results}
After we completed the new test set, we evaluated a broad range of image classification models.
The main question was how the accuracy on the original CIFAR-10 test set compares to the accuracy on our new test set.
To this end, we experimented with a broad range of classifiers spanning multiple years of machine learning research.
The models include widely used convolutional networks (VGG and ResNet \cite{vgg,resnet}), more recent architectures (ResNeXt, PyramidNet, DenseNet \cite{resnext,pyramidnet,densenet}), the published state-of-the-art (Shake-Drop \cite{shakedrop}), and a model derived from RL-based hyperparameter search (NASNet) \cite{nas}.
In addition, we also evaluated ``shallow'' approaches based on random features \cite{rahimi2009weighted,rf}.
Overall, the accuracies on the original CIFAR-10 test set range from about 80\% to 97\%.

For all deep architectures, we used code previously published online (see Appendix \ref{apx:model_descriptions} for a list).
To avoid bias due to specific model repositories or frameworks, we also evaluated two widely used architectures (VGG and ResNets) from two different sources implemented in different deep learning libraries.
We wrote our own implementation for the models based on random features.

Our main results are summarized in Table \ref{tab:v4_results} and Figure \ref{fig:regression_plot}.
We now describe the two important trends here and then discuss our results in Section \ref{sec:conclusion}.

\subsection{A Significant Drop in Accuracy}
All models see a large drop in accuracy from the original to the new test set.
The \emph{absolute} gap is larger for models that perform worse on the original test set and smaller for models with better published CIFAR-10 accuracy.
For instance, VGG and ResNet architectures see a gap of about 8\% between their original accuracy (around 93\%) and their new accuracy (around 85\%).
The best original accuracy is achieved by \model{shake\_shake\_64d\_cutout}, which sees a roughly 4\% drop from 97\% to 93\%.
While there is some variation in the accuracy drop, no model is a clear outlier.

In terms of \emph{relative} error, the models with higher original accuracy tend to have a larger increase.
Some of the models such as \model{DARC}, \model{shake\_shake\_32d}, and \model{resnext\_29\_4x64d} see a $3\times$ increase in their error rate.
For simpler models such as VGG, AlexNet, or ResNet, the relative error increase is in the range $1.7\times$ to $2.3\times$.
We refer the reader to Appendix \ref{apx:v4_error_results} for a table with all relative error numbers.

\subsection{Few Changes in the Relative Order}
When sorting the models in order of their original and new accuracy, there are few changes in the overall ranking.
Models with comparable original accuracy tend to see a similar decrease in performance.
In fact, Figure \ref{fig:regression_plot} shows that the relationship between original and new accuracy can be explained well with a linear function derived from a least squares fit.
The new accuracy of a model is roughly given by the following formula:
\begin{equation}
  \accnew \; = \; (1.62\pm0.04) \cdot \accorig - 65.51\% \pm3.16\% \; .
\end{equation}

On the other hand, it is worth noting that some techniques give a consistently larger increase on the new test set.
For instance, adding the Cutout data augmentation \cite{cutout} to a \model{shake\_shake\_64d} network adds only 0.12\% accuracy on the original test set but gives an accuracy increase of about 1.5\% on the new test set. 
Similarly, adding Cutout to a \model{wide\_resnet\_28\_10} classifiers improves the accuracy by about 1\% on the original test set and 2.2\% on the new test set.
As another example, note that increasing the \emph{width} of a ResNet as opposed to its \emph{depth} provides larger benefits on the new test set.
 
\begin{table}
  \caption{Model accuracy on the original CIFAR-10 test set and the new test set, with the gap reported as the difference between the two accuracies.
$\Delta$ Rank is the relative difference in the ranking from the original test set to the new test
set.
For example, $\Delta \text{Rank} = -2$ means a model dropped in the rankings by two positions
on the new test set. 
}
  \label{tab:v4_results}
  \centering
  \rowcolors{2}{white}{gray!25}
  \begin{tabular}{lrrrr}
\toprule
{} &                                    Original Accuracy &                                         New Accuracy &  Gap & $\Delta$ Rank \\
\midrule
\model{shake\_shake\_64d\_cutout}   \cite{shakeshake, cutout} &  97.1 {\footnotesize \textcolor{gray}{[96.8, 97.4]}} &  93.0 {\footnotesize \textcolor{gray}{[91.8, 94.0]}} &  4.1 &  0 \\
\model{shake\_shake\_96d}   \cite{shakeshake}                 &  97.1 {\footnotesize \textcolor{gray}{[96.7, 97.4]}} &  91.9 {\footnotesize \textcolor{gray}{[90.7, 93.1]}} &  5.1 &  -2 \\
\model{shake\_shake\_64d}   \cite{shakeshake}                 &  97.0 {\footnotesize \textcolor{gray}{[96.6, 97.3]}} &  91.4 {\footnotesize \textcolor{gray}{[90.1, 92.6]}} &  5.6 &  -2 \\
\model{wide\_resnet\_28\_10\_cutout}   \cite{wrn, cutout}     &  97.0 {\footnotesize \textcolor{gray}{[96.6, 97.3]}} &  92.0 {\footnotesize \textcolor{gray}{[90.7, 93.1]}} &  5 &  +1 \\
\model{shake\_drop}   \cite{shakedrop}                        &  96.9 {\footnotesize \textcolor{gray}{[96.5, 97.2]}} &  92.3 {\footnotesize \textcolor{gray}{[91.0, 93.4]}} &  4.6 &  +3 \\
\model{shake\_shake\_32d}   \cite{shakeshake}                 &  96.6 {\footnotesize \textcolor{gray}{[96.2, 96.9]}} &  89.8 {\footnotesize \textcolor{gray}{[88.4, 91.1]}} &  6.8 &  -2 \\
\model{darc}   \cite{darc}                                    &  96.6 {\footnotesize \textcolor{gray}{[96.2, 96.9]}} &  89.5 {\footnotesize \textcolor{gray}{[88.1, 90.8]}} &  7.1 &  -4 \\
\model{resnext\_29\_4x64d}   \cite{resnext}                   &  96.4 {\footnotesize \textcolor{gray}{[96.0, 96.7]}} &  89.6 {\footnotesize \textcolor{gray}{[88.2, 90.9]}} &  6.8 &  -2 \\
\model{pyramidnet\_basic\_110\_270}   \cite{pyramidnet}       &  96.3 {\footnotesize \textcolor{gray}{[96.0, 96.7]}} &  90.5 {\footnotesize \textcolor{gray}{[89.1, 91.7]}} &  5.9 &  +3 \\
\model{resnext\_29\_8x64d}   \cite{resnext}                   &  96.2 {\footnotesize \textcolor{gray}{[95.8, 96.6]}} &  90.0 {\footnotesize \textcolor{gray}{[88.6, 91.2]}} &  6.3 &  +3 \\
\model{wide\_resnet\_28\_10}   \cite{wrn}                     &  95.9 {\footnotesize \textcolor{gray}{[95.5, 96.3]}} &  89.7 {\footnotesize \textcolor{gray}{[88.3, 91.0]}} &  6.2 &  +2 \\
\model{pyramidnet\_basic\_110\_84}   \cite{pyramidnet}        &  95.7 {\footnotesize \textcolor{gray}{[95.3, 96.1]}} &  89.3 {\footnotesize \textcolor{gray}{[87.8, 90.6]}} &  6.5 &  0 \\
\model{densenet\_BC\_100\_12}   \cite{densenet}               &  95.5 {\footnotesize \textcolor{gray}{[95.1, 95.9]}} &  87.6 {\footnotesize \textcolor{gray}{[86.1, 89.0]}} &  8 &  -2 \\
\model{neural\_architecture\_search}   \cite{nas}             &  95.4 {\footnotesize \textcolor{gray}{[95.0, 95.8]}} &  88.8 {\footnotesize \textcolor{gray}{[87.4, 90.2]}} &  6.6 &  +1 \\
\model{wide\_resnet\_tf}   \cite{wrn}                         &  95.0 {\footnotesize \textcolor{gray}{[94.6, 95.4]}} &  88.5 {\footnotesize \textcolor{gray}{[87.0, 89.9]}} &  6.5 &  +1 \\
\model{resnet\_v2\_bottleneck\_164}   \cite{resnet_preact}    &  94.2 {\footnotesize \textcolor{gray}{[93.7, 94.6]}} &  85.9 {\footnotesize \textcolor{gray}{[84.3, 87.4]}} &  8.3 &  -1 \\
\model{vgg16\_keras}   \cite{vgg, vgg_cifar}                  &  93.6 {\footnotesize \textcolor{gray}{[93.1, 94.1]}} &  85.3 {\footnotesize \textcolor{gray}{[83.6, 86.8]}} &  8.3 &  -1 \\
\model{resnet\_basic\_110}   \cite{resnet}                    &  93.5 {\footnotesize \textcolor{gray}{[93.0, 93.9]}} &  85.2 {\footnotesize \textcolor{gray}{[83.5, 86.7]}} &  8.3 &  -1 \\
\model{resnet\_v2\_basic\_110}   \cite{resnet_preact}         &  93.4 {\footnotesize \textcolor{gray}{[92.9, 93.9]}} &  86.5 {\footnotesize \textcolor{gray}{[84.9, 88.0]}} &  6.9 &  +3 \\
\model{resnet\_basic\_56}   \cite{resnet}                     &  93.3 {\footnotesize \textcolor{gray}{[92.8, 93.8]}} &  85.0 {\footnotesize \textcolor{gray}{[83.3, 86.5]}} &  8.3 &  0 \\
\model{resnet\_basic\_44}   \cite{resnet}                     &  93.0 {\footnotesize \textcolor{gray}{[92.5, 93.5]}} &  84.2 {\footnotesize \textcolor{gray}{[82.6, 85.8]}} &  8.8 &  -3 \\
\model{vgg\_15\_BN\_64}   \cite{vgg, vgg_cifar}               &  93.0 {\footnotesize \textcolor{gray}{[92.5, 93.5]}} &  84.9 {\footnotesize \textcolor{gray}{[83.2, 86.4]}} &  8.1 &  +1 \\
\model{resnet\_preact\_tf}   \cite{resnet}                    &  92.7 {\footnotesize \textcolor{gray}{[92.2, 93.2]}} &  84.4 {\footnotesize \textcolor{gray}{[82.7, 85.9]}} &  8.3 &  0 \\
\model{resnet\_basic\_32}   \cite{resnet}                     &  92.5 {\footnotesize \textcolor{gray}{[92.0, 93.0]}} &  84.9 {\footnotesize \textcolor{gray}{[83.2, 86.4]}} &  7.7 &  +3 \\
\model{cudaconvnet}   \cite{alexnet}                          &  88.5 {\footnotesize \textcolor{gray}{[87.9, 89.2]}} &  77.5 {\footnotesize \textcolor{gray}{[75.7, 79.3]}} &  11 &  0 \\
\model{random\_features\_256k\_aug}   \cite{rf}               &  85.6 {\footnotesize \textcolor{gray}{[84.9, 86.3]}} &  73.1 {\footnotesize \textcolor{gray}{[71.1, 75.1]}} &  12 &  0 \\
\model{random\_features\_32k\_aug}   \cite{rf}                &  85.0 {\footnotesize \textcolor{gray}{[84.3, 85.7]}} &  71.9 {\footnotesize \textcolor{gray}{[69.9, 73.9]}} &  13 &  0 \\
\model{random\_features\_256k}   \cite{rf}                    &  84.2 {\footnotesize \textcolor{gray}{[83.5, 84.9]}} &  69.9 {\footnotesize \textcolor{gray}{[67.8, 71.9]}} &  14 &  0 \\
\model{random\_features\_32k}   \cite{rf}                     &  83.3 {\footnotesize \textcolor{gray}{[82.6, 84.0]}} &  67.9 {\footnotesize \textcolor{gray}{[65.9, 70.0]}} &  15 &  -1 \\
\model{alexnet\_tf}                                           &  82.0 {\footnotesize \textcolor{gray}{[81.2, 82.7]}} &  68.9 {\footnotesize \textcolor{gray}{[66.8, 70.9]}} &  13 &  +1 \\
\bottomrule
\end{tabular}

\end{table}

\begin{figure}[h!]
\centering
\begin{subfigure}{.45\textwidth}
	\includegraphics[width=\linewidth]{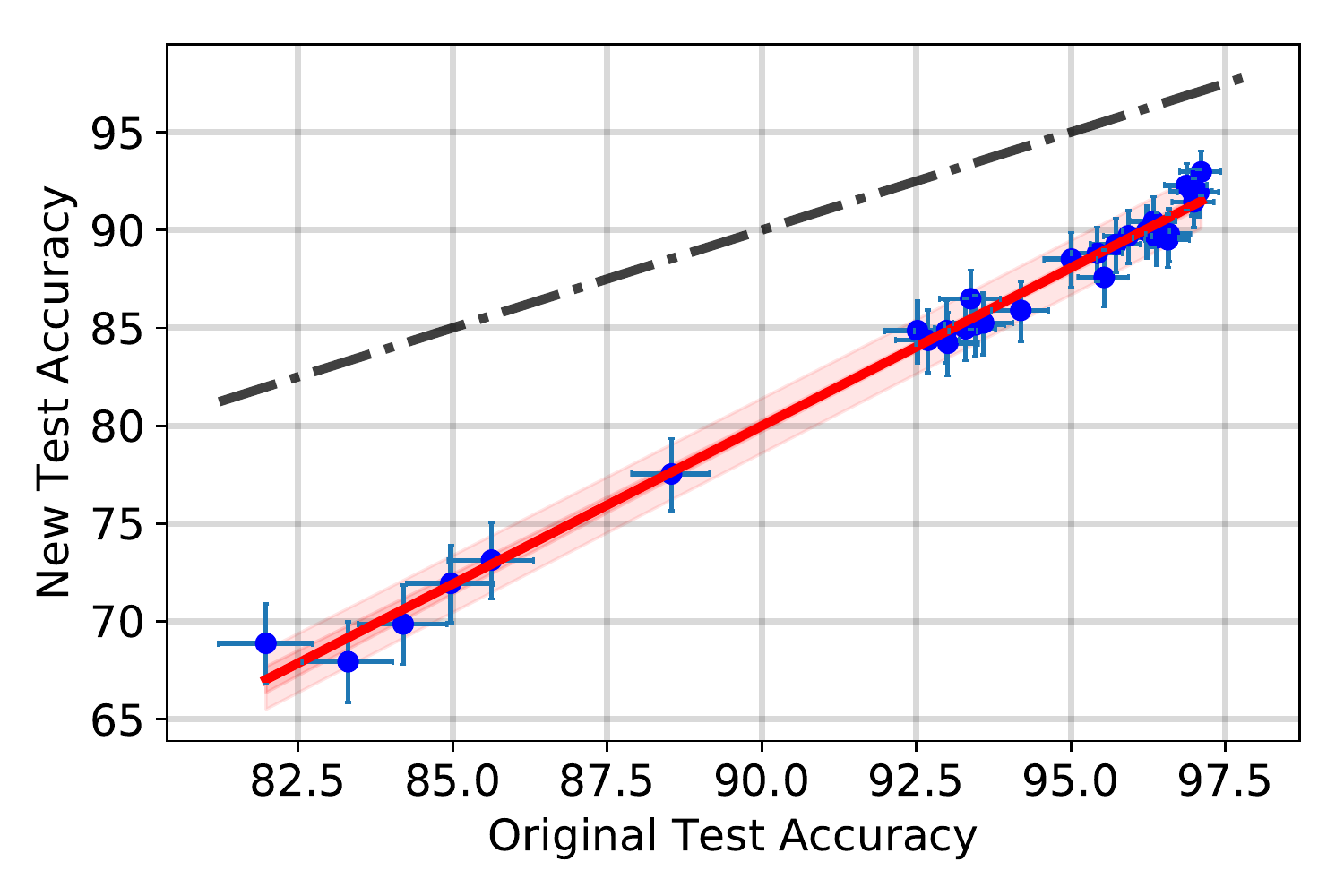}
\caption{All models}\label{subfig:all_models}
\end{subfigure}
\begin{subfigure}{0.45\textwidth}
	\includegraphics[width=\linewidth]{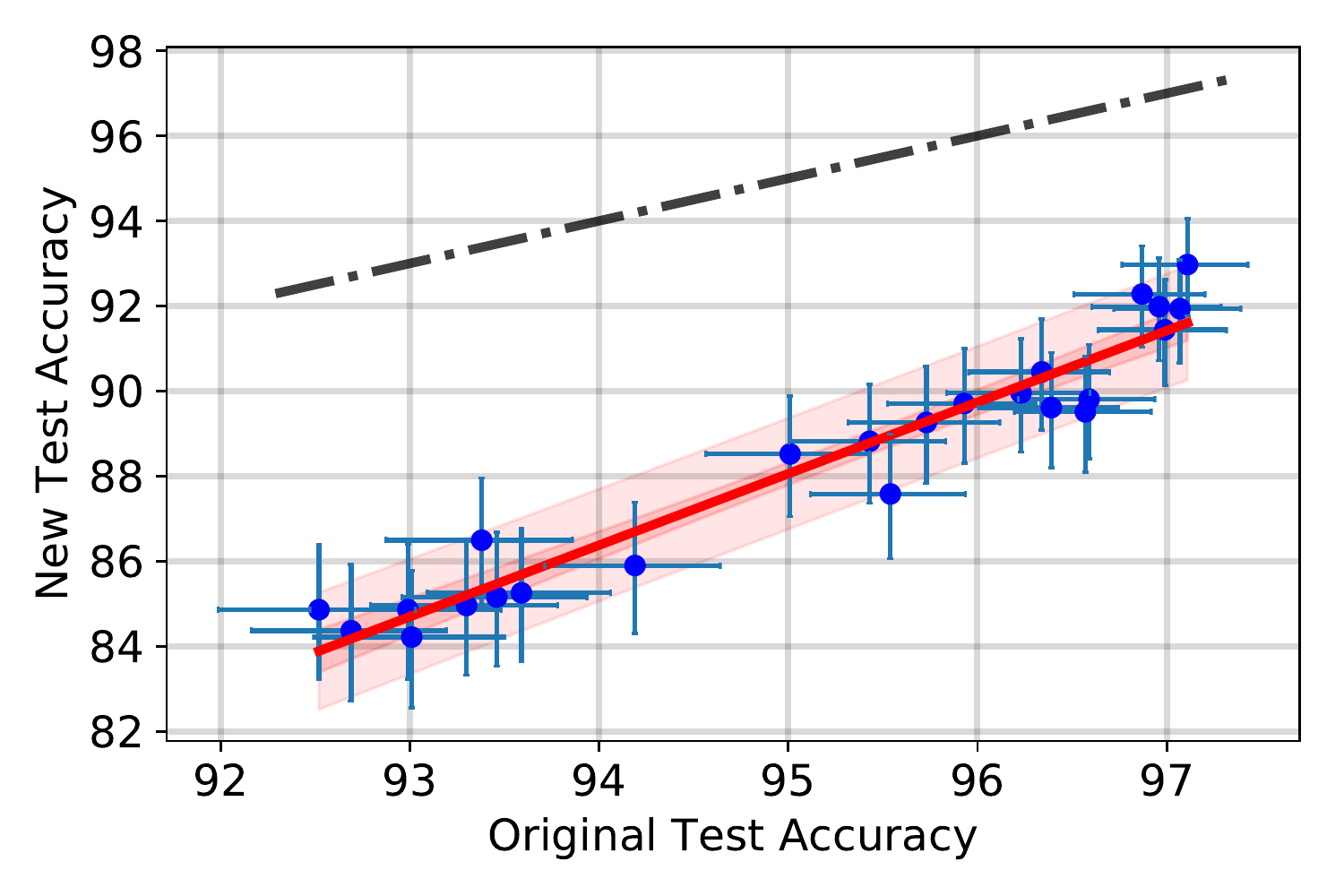}
 \caption{High accuracy models}\label{subfig:high_accuracy_models}
\end{subfigure} \\
\begin{subfigure}{\textwidth}
	\includegraphics[width=\linewidth]{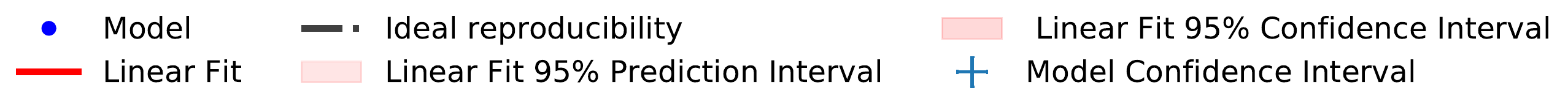}
\end{subfigure}
\caption{Model accuracy on new test set vs.\ model accuracy on original test set.}
\label{fig:regression_plot}
\end{figure}

\iftoggle{isnips}{}{
\subsection{A Model for the Linear Fit}
\label{sec:linear_fit}
Though the linear fit observed in Figure \ref{fig:regression_plot} rules out that the new test set is identically distributed as the original test set, the linear relationship between the old and new test errors is striking. There are a variety of plausible explanations for this effect. For instance, posit that the original test set is composed of two sub-populations.
On the ``easy'' sub-population, a classifier achieves an accuracy of $a_0$.
The ``hard'' sub-population is $\kappa$ times more difficult in the sense that the classification error on these examples is $\kappa$ times larger.
Hence the accuracy on this sub-population is $1 - \kappa (1 - a_0)$.
If the relative frequencies of these two sub-populations are $p_1$ and $p_2$, we get the following overall accuracy:
\[
  \accorig \; = \; p_1 \cdot a_0 + p_2 \cdot (1 - \kappa (1 - a_0))
\]
which we can rewrite as a simple linear function of $a_0$:
\[
  \accorig \; = \; \beta \cdot a_0 + \gamma \; .
\]

For the new test set, we also assume a mixture distribution consisting of a different proportion of the same two components, with relative frequencies now $q_1$ and $q_2$.
We can then write the accuracy on the new test set as
\begin{align*}
  \accnew \; &= \; q_1 \cdot a_0 + q_2 \cdot (1 - \kappa (1 - a_0)) \\
             &= \; \beta' \cdot a_0 + \gamma'
\end{align*}
where we collected terms into a simple linear function as before.

It is now easy to see that the new accuracy is indeed a linear function of the original accuracy:
\begin{align*}
\accnew \; &= \; \frac{\beta'}{\beta} (\beta a_0 + \gamma) - \frac{\beta'}{\beta} \gamma + \gamma'\\
          &= \; \frac{\beta'}{\beta} \accorig  \; .
\end{align*}

We remark that we do not see this mixture model as a ground truth explanation, but rather as an illustrative example for how a linear dependency between the original and new test accuracies naturally arises with small distribution shifts between data sets. In reality, the two test sets have a more complex composition with different accuracies on various sub-populations.
Nevertheless, this model reveals surprising sensitivities can exist from distribution shift even while relative ordering of classifiers remain constant. We hope that such sensitivities to distribution shift can be experimentally validated in future work.
}

\section{Explaining the Gap}
\iftoggle{isnips}{
\input{explain_gap}
}{
Since the gap between original and new accuracy is concerningly large, we investigated multiple hypotheses for explaining this gap.

\subsection{Statistical error}
A first natural guess is that the gap is simply due to statistical fluctuations.
But as noted before, the sample size of our new test set is large enough so that a 95\% confidence interval has size about $\pm 1.2\%$.
Since a 95\% confidence interval for the original CIFAR-10 test accuracy is even smaller (roughly $\pm 0.6\%$ for 90\% classification accuracy and $\pm 0.3\%$ for 97\% classification accuracy), we can rule out statistical error as the sole explanation.

\subsection{Differences in near-duplicate removal}
As mentioned in Section \ref{sec:building_new_test_set}, the final step of both the original CIFAR-10 and our dataset creation procedure is a near-duplicate removal.
While removing near-duplicates between our new test set and the original CIFAR-10 dataset, we noticed that the latter contained images that we would have ruled out as near-duplicates.
A large number of near-duplicates between CIFAR-10 train and test, combined with our more stringent near-duplicate removal, could explain some of the accuracy drop.
Indeed, we found about 800 images in the CIFAR-10 test set that we would classify as near-duplicates.
Moreover, most classifiers have accuracy between 99\% and 100\% on these near-duplicates (recall that most models achieve 100\% training error).
But since the 800 images comprise only 8\% of the original test set, the near-duplicates can explain at most 1\% of the observed difference.

For completeness, we describe our process for finding near duplicates in detail.
For every test image, we visually inspected the top-10 nearest neighbors in both $\ell_2$-distance and the SSIM (structural similarity) metric.
We compared the original test set to the CIFAR-10 training set, and our new test set to both the original training and test sets.
We consider an image pair as near-duplicates if both images have the same object in the same pose. 
We include images that have different zoom, color scale, stretch in the horizontal or vertical direction, or small shifts in vertical or horizontal position.  
If the object was rotated or in a different pose, we did not include it as a near-duplicate.

\subsection{Hyperparameter tuning}
Another conjecture is that we can recover some of the missing accuracy by re-tuning hyperparameters of a model.
To this end, we performed a grid search over multiple parameters of a VGG model.
We selected three standard hyperparameters known to strongly influence test set performance: initial learning rate, dropout, and weight decay. 
The \model{vgg16\_keras} architecture uses different amounts of dropout across different layers of the network, so we chose to tune a multiplicative scaling factor for the amount of dropout, keeping the ratio of dropout across different layers constant. 

We initialized a hyperparameter configuration from values tuned to the original test set (learning
rate = $0.1$, dropout ratio = $1$, weight decay = $5\mathrm{e}-4$), and performed a
grid search across the following values:
\begin{itemize}
\item learning rate in $ \{0.0125, 0.025, 0.05, 0.1, 0.2, 0.4, 0.8\}$
\item dropout ratio in $\{0.5, 0.75, 1, 1.25, 1.75\}$ 
\item weight decay in $\{5\mathrm{e}{-5}, 1\mathrm{e}{-4}, 5\mathrm{e}{-4}, 1\mathrm{e}{-3},
5\mathrm{e}{-3}\}$
\end{itemize}

We ensured that the best performance was never at an extreme point of any of the ranges we tested for an individual hyperparameter. 
However, we did not find a setting with a significantly better accuracy on the new test set (the biggest improvement was from 85.25\% to 85.84\%).

\subsection{Inspecting hard images}
It is also possible that we accidentally created a more difficult test set by including a set of "harder" images.  
To explore this, we visually inspected the set of images that the majority of models incorrectly classified. 
We find that all the new images are natural images that are recognizable to humans.
Figure \ref{fig:hardtest} in Appendix \ref{apx:explain_gap} shows examples of the hard images in our new test set that no model correctly classified.

\subsection{Training on part of our new test set}
If our new test set came from a significantly different data distribution than the original CIFAR-10 dataset, then retraining on half of our new test set plus the original training set should improve the accuracy scores on the held-out fraction of the new test set.  

We conducted this experiment by randomly drawing a class-balanced split containing 1010 images from the new test set.
We then added these images to the full CIFAR-10 training set and retrained the \model{vgg16\_keras} model.  After training, we tested the model on the 1011 held-out images from the new test set.
We repeated this experiment twice with different randomly selected splits from our test set, obtaining accuracies of 85.06\% and 85.36\% (compared to 84.9\% without the extra training data).
This provides further evidence that there are no large distribution shifts between our new test set and the original CIFAR-10 dataset. 

\subsection{Cross-validation}
Since cross-validation is a more principled way of measuring a model's generalization ability, we tested if cross-validation on the original CIFAR-10 dataset could predict a model's error on our new test set. 
We created cross-validation data by randomly dividing the training set into 5 class-balanced splits.
We then randomly shuffled together 4 out of the 5 training splits with toe original test set.
The leftover held-out split from the training set then became the new test set.

We retrained the models \model{vgg\_15\_BN\_64}, \model{wide\_resnet\_28\_10}, and \model{shake\_shake\_64d\_cutout} on each of the 5 new datasets we created.  
The accuracies are reported in Table \ref{tab:cross_validation}.
The accuracies on each of the cross validation splits did not vary much from the accuracies on the original test set.  

\begin{table}[h!]
\centering
\begin{tabular}{c | c | c | c}
 & \model{vgg\_15\_BN\_64} & \model{wide\_resnet\_28\_10} & \model{shake\_shake\_64d\_cutout} \\ \hline
Split 1 & 93.87 & 96.16 &	97.16 \\
Split 2 & 93.81	& 96.04 &	97.3\\
Split 3 & 94.01	& 96.37 &	97.38\\
Split 4 & 93.99	& 96.16 &	97.39\\
Split 5 & 93.5	& 96.5  & 	97.4\\
\end{tabular}
\caption{Model Accuracies on cross validation splits}
\label{tab:cross_validation}
\end{table}

}


\section{Discussion}
\label{sec:conclusion}
\paragraph{Overfitting.} Do our experiments reveal overfitting?
This is arguably the main question when interpreting our results.
To be precise, we first define two notions of overfitting:

\iftoggle{isnips}{
\begin{itemize}[leftmargin=1cm]
}{
\begin{itemize}
}
\item \textbf{Training set overfitting.} One way to quantify overfitting is as the difference between the training accuracy and the test accuracy.
Note that the deep neural networks in our experiments usually achieve 100\% training accuracy.
So this notion of overfitting already occurs on the existing  dataset.

\item \textbf{Test set overfitting.} Another notion of overfitting is the gap between the test accuracy and the accuracy on the underlying data distribution.
By adapting model design choices to the test set, the concern is that we implicitly fit the model to the test set.
The test accuracy then loses its validity as an accurate measure of performance on truly unseen data.
\end{itemize}

Since the overall goal in machine learning is to generalize to unseen data, we argue that the second notion of overfitting through test set adaptivity is more important.
Surprisingly, our results show no signs of such overfitting on CIFAR-10.
Despite multiple years of competitive adaptivity on this dataset, there has been no stagnation on truly held out data.
In fact, the best performing models on our new test set see an \emph{increased} advantage over more established baselines.
Though this trend is opposite to what overfitting through adaptivity would suggest. While a conclusive picture will require further replication experiments, we view our results as support of the competition-based approach to increasing accuracy scores.

We note that one can read the analysis of Blum and Hardt's Ladder algorithm as supporting this claim~\cite{BH15}. Indeed, they show that adding a minor modification of standard machine learning competitions avoids the sort of overfitting that can be achieved with aggressive adaptivity. Our results show that even without these modifications, model tuning based on the test error does not lead to overfitting on a standard data set.

\paragraph{Distribution shift.}
Although our results do not support the hypothesis of adaptivity-based overfitting, there is still a significant gap between original and new accuracy scores that needs to be explained.
We view this gap as the result of a small distribution shift between the original CIFAR-10 dataset and our new test set.
The fact that this gap is large, affects all models, and occurs despite our efforts to replicate the CIFAR-10 creation process is concerning.
Normally, distribution shift is studied for specific changes in the data generation process (e.g., changes in lighting conditions) or for worst-case attacks in an adversarial setting.
Our experiment is more benign and poses neither of these challenges.
Nevertheless, the accuracy of all models drops by 4 - 15\% and the relative increase in error rates is up to $3 \times$.
This indicates that current CIFAR-10 classifiers have difficulty generalizing to natural variations in image data.

\paragraph{Future work.}
Concrete future experiments should explore whether the competition approach is similarly resilient to overfitting on other datasets (e.g., ImageNet) and other tasks (such as language modeling).
An important aspect here is to ensure that the data distribution of a new test set stays as close to the original dataset as possible.
Furthermore, we should understand what types of naturally occurring distribution shifts are challenging for image classifiers.
For instance, are there certain sub-populations that the models fail to learn on CIFAR-10 but appear
trivial to a human? 
\iftoggle{isnips}{
In Appendix \ref{sec:linear_fit}, we describe a simple mixture model based on sub-population shifts that also exhibits the linear trend observed in Figure \ref{fig:regression_plot}.}{
In Section \ref{sec:linear_fit}, we described a simple mixture model based on sub-population shifts that could serve as a starting point for such an investigation.}

More broadly, we view our results as motivation for a more thorough evaluation of machine learning research.
Currently, the dominant paradigm is to propose a new algorithm and evaluate its performance on existing data.
Unfortunately, there is often little understanding to what extent the improvements are broadly applicable.
To truly understand \emph{generalization} questions, more studies should collect insightful new data and evaluate existing algorithms on such data.
Since we now have a large body of essentially pre-registered classifiers in open-source repositories, such studies would conform to well established standards of statistically valid research.
It is important to note the distinction to current reproducibility efforts in machine learning that usually focus on \emph{computational} reproducibility, i.e., running published code on the same test data.
In contrast, generalization experiments such as ours focus on \emph{statistical} reproducibility by evaluating classifiers on truly new data (similar to recruiting new participants for a reproducibility experiment in medicine or psychology).

\section*{Acknowledgements}
Benjamin Recht and Vaishaal Shankar are supported by ONR award N00014-17-1-2502.
Benjamin Recht is additionally generously supported in part by NSF award CCF-1359814, ONR awards N00014-17-1-219 and N00014-17-1-2401, the DARPA Fundamental Limits of Learning (Fun LoL) and Lagrange Programs, and an Amazon AWS AI Research Award.
Rebecca Roelofs is generously supported by DOE award AC02-05CH11231.
Ludwig Schmidt is generously supported by a Google PhD fellowship.
Ludwig was a visitor at UC Berkeley while conducting the research for this paper.

We would like to thank Alexei Efros, David Fouhey, and Moritz Hardt for helpful discussions while working on this paper.

\bibliographystyle{plainnat}
\bibliography{main}

\newpage

\appendix
\iftoggle{isnips}{
\section{A Model for the Linear Fit}
\label{sec:linear_fit}
}{}

\section{Model Descriptions}
\label{apx:model_descriptions}

\begin{table}[h!]
\centering
\begin{tabular}{ll}
\toprule
{} &                                     Code Repository \\
\midrule
\model{alexnet\_tf}                                           &  \url{} \\
\model{cudaconvnet}   \cite{alexnet}                          &  \url{akrizhevsky/cuda-convnet2} \\
\model{darc}   \cite{darc}                                    &  \url{http://lis.csail.mit.edu/code/gdl.html} \\
\model{densenet\_BC\_100\_12}   \cite{densenet}               &  \url{hysts/pytorch\_image\_classification/} \\
\model{pyramidnet\_basic\_110\_270}   \cite{pyramidnet}       &  \url{hysts/pytorch\_image\_classification/} \\
\model{pyramidnet\_basic\_110\_84}   \cite{pyramidnet}        &  \url{hysts/pytorch\_image\_classification/} \\
\model{resnet\_basic\_110}   \cite{resnet}                    &  \url{hysts/pytorch\_image\_classification/} \\
\model{resnet\_basic\_32}   \cite{resnet}                     &  \url{hysts/pytorch\_image\_classification/} \\
\model{resnet\_basic\_44}   \cite{resnet}                     &  \url{hysts/pytorch\_image\_classification/} \\
\model{resnet\_basic\_56}   \cite{resnet}                     &  \url{hysts/pytorch\_image\_classification/} \\
\model{resnet\_preact\_tf}   \cite{resnet}                    &  \url{tensorflow/models/../slim/nets/resnet\_v2.py} \\
\model{resnet\_v2\_basic\_110}   \cite{resnet_preact}         &  \url{hysts/pytorch\_image\_classification/} \\
\model{resnet\_v2\_bottleneck\_164}   \cite{resnet_preact}    &  \url{hysts/pytorch\_image\_classification/} \\
\model{resnext\_29\_4x64d}   \cite{resnext}                   &  \url{hysts/pytorch\_image\_classification/} \\
\model{resnext\_29\_8x64d}   \cite{resnext}                   &  \url{hysts/pytorch\_image\_classification/} \\
\model{shake\_drop}   \cite{shakedrop}                        &  \url{imenurok/ShakeDrop} \\
\model{shake\_shake\_32d}   \cite{shakeshake}                 &  \url{hysts/pytorch\_image\_classification/} \\
\model{shake\_shake\_64d\_cutout}   \cite{shakeshake, cutout} &  \url{hysts/pytorch\_image\_classification/} \\
\model{shake\_shake\_64d}   \cite{shakeshake}                 &  \url{hysts/pytorch\_image\_classification/} \\
\model{shake\_shake\_96d}   \cite{shakeshake}                 &  \url{hysts/pytorch\_image\_classification/} \\
\model{vgg16\_keras}   \cite{vgg, vgg_cifar}                  &  \url{geifmany/cifar-vgg} \\
\model{vgg\_15\_BN\_64}   \cite{vgg, vgg_cifar}               &  \url{hysts/pytorch\_image\_classification/} \\
\model{wide\_resnet\_28\_10\_cutout}   \cite{wrn, cutout}     &  \url{hysts/pytorch\_image\_classification/} \\
\model{wide\_resnet\_28\_10}   \cite{wrn}                     &  \url{hysts/pytorch\_image\_classification/} \\
\model{wide\_resnet\_tf}   \cite{wrn}                         &  \url{tensorflow/models/tree/master/resnet} \\
\bottomrule
\end{tabular}

\caption{Code repositories for deep models. With the exception of \model{darc}, repositories are hosted at
\url{https://github.com/}}
\label{tab:model_code_repos}
\end{table}

Table \ref{tab:model_code_repos} contains the code repositories for the deep model.  
The specific parameters we list below are the default configurations for each model in the code repositories. Unless otherwise noted, all models were trained on a single GPU. 
\begin{itemize}
  \item \model{alexnet\_tf} \cite{alexnet} : Alexnet model without data augmentation
\item \model{cudaconvnet}   \cite{alexnet}: Alexnet model with data augmentation
\item \model{darc}   \cite{darc}: A ResNeXt \cite{resnext} with depth 29 and an extra regularizer for promoting generalization
\item \model{densenet\_BC\_100\_12}:  DenseNet \cite{densenet} with batch size 32 , initial learning rate 0.05, depth 100, block type ``bottleneck", growth rate 12, compression rate 0.5
\item \model{nas}:   Neural Architecture Search \cite{nas} model for CIFAR-10, batch size 32, learning rate 0.025, cosine (single period) learning rate decay, auxiliary head loss weighting 0.4, clip global norm of all gradients by 5
\item \model{pyramidnet\_basic\_110\_270}:  PyramidNet \cite{pyramidnet} with depth 110, block type ``basic", $\alpha=270$
\item \model{pyramidnet\_basic\_110\_84}:  PyramidNet \cite{pyramidnet} with depth 110, block type ``basic", $\alpha=84$
\item \model{random\_features\_256k\_aug}   \cite{rf}: Random 1 layer convolutional network with 256k filters sampled from image patches, patch size = $6$, pool size $15$, pool stride $16$, and horizontal flip data augmentation.
\item \model{random\_features\_256k}   \cite{rf}: Random 1 layer convolutional network with 256k filters sampled from image patches, patch size = $6$, pool size $15$, pool stride $16$.
\
\item \model{random\_features\_32k\_aug}   \cite{rf}: Random 1 layer convolutional network with 32k filters sampled from image patches, patch size = $6$, pool size $15$, pool stride $16$, and horizontal flip data augmentation.
\
\item \model{random\_features\_32k}   \cite{rf}: Random 1 layer convolutional network with 32k filters sampled from image patches, patch size = $6$, pool size $15$, pool stride $16$
\
\item \model{resnet\_basic\_32}: ResNet  \cite{resnet} with depth 32, block type ``basic"
\item \model{resnet\_basic\_44}: ResNet  \cite{resnet} with depth 44, block type ``basic"
\item \model{resnet\_basic\_56}: ResNet  \cite{resnet} with depth 56, block type ``basic"
\item \model{resnet\_basic\_110}: ResNet  \cite{resnet} with depth 110, block type ``basic"
\item \model{resnet\_preact\_basic\_110}: ResNet-preact  \cite{resnet_preact} with depth 110, block type ``basic"
\item \model{resnet\_preact\_bottleneck\_164}: ResNet-preact \cite{resnet_preact} with depth 164, block type ``bottleneck" 
\item \model{resnetv2\_tf} : ResNet-preact  \cite{resnet_preact} with depth 32, block type ``basic" (implemented in TensorFlow)
\item \model{resnext\_29\_4x64d}:  ResNeXt \cite{resnext} with depth 29, cardinality 4, base channels 64, batch size 32 and initial learning rate 0.025 
\item \model{resnext\_29\_8x64d}:  ResNeXt \cite{resnext} with depth 29, cardinality 8, base channels 64, batch size 64 and initial learning rate 0.05 
\item \model{shake\_drop}: Shake-drop \cite{shakedrop} with PyramidNet, epochs 300, batch size 128, lr 0.5, shared gradient input, 4 GPUS
\item \model{shake\_shake\_32d} : Shake-shake \cite{shakeshake} with depth 26, base channels 32, S-S-I model
\item \model{shake\_shake\_64d}: Shake-shake \cite{shakeshake} with depth 26, base channels 64, S-S-I model, batch size 64, base lr 0.1
\item \model{shake\_shake\_96d}:  Shake-shake \cite{shakeshake} with depth 26, base channels 96, S-S-I model
\item \model{shake\_shake\_64d\_cutout}: Shake-shake  \cite{shakeshake}  with depth 26, base channels 64, S-S-I model,  batch size 64, lr 0.1, cosine scheduler, cutout \cite{cutout} size 16
\item \model{vgg16\_keras}:  VGG \cite{vgg, vgg_cifar} with depth 16, batch normalization
\item \model{vgg\_15\_BN\_64} : VGG  \cite{vgg, vgg_cifar} with depth 15, batch normalization, 64 channels
\item \model{wide\_resnet\_tf} : Wide residual network \cite{wrn} with widening factor 10 (implemented in TensorFlow)
\item \model{wide\_resnet\_28\_10} :  Wide residual network \cite{wrn} with depth 28, widening factor 10
\item \model{wide\_resnet\_28\_10\_cutout} : Wide residual network \cite{wrn} with depth 28, widening factor 10, base lr 0.1, batch size 64, cosine scheduler, cutout \cite{cutout} size 16
\end{itemize}

\section{Details for ``Explaining the Gap" Experiments}
\label{apx:explain_gap}

\iftoggle{isnips}{
\paragraph{Statistical Error.}
To obtain the 95\% confidence intervals, we used a Clopper-Pearson interval, which is a conservative confidence interval for a binomial distribution of $k$ expected successes on $n$ trials. 

\paragraph{Differences in near-duplicate removal.}
We found near duplicates by visually inspecting the top-10 nearest neighbors in the training set for each image in the original test set.  
We performed this check using both l2 and structural similarity (SSIM) distance metrics, finding that SSIM allowed us to detect slightly more near duplicates. 
We consider an image pair as near-duplicates if both images have the same object in the same pose. 
We include images that have different zoom, color scale, stretch in the horizontal or vertical direction, or small shifts in vertical or horizontal position.  
If the object was rotated or in a different pose, we did not include it as a near-duplicate.

\paragraph{Hyperparameter tuning.}
We focused on the VGG architecture and selected three standard hyperparameters known to strongly influence test set performance: initial learning rate, dropout, and weight decay. 
The \model{vgg16\_keras} architecture uses different amounts of dropout across different layers of the network, so we chose to tune a multiplicative scaling factor for the amount of dropout, keeping the ratio of dropout across different layers constant. 
We initialized a hyperparameter configuration from values tuned to the original test set (learning
rate = $0.1$, dropout ratio = $1$, weight decay = $5\mathrm{e}-4$), and performed a
grid search across the following values:
\begin{itemize}
\item learning rate in $ \{0.0125, 0.025, 0.05, 0.1, 0.2, 0.4, 0.8\}$
\item dropout ratio in $\{0.5, 0.75, 1, 1.25, 1.75\}$ 
\item weight decay in $\{5\mathrm{e}{-5}, 1\mathrm{e}{-4}, 5\mathrm{e}{-4}, 1\mathrm{e}{-3},
5\mathrm{e}{-3}\}$
\end{itemize}

We ensured that the best performance was never at an extreme point of any of the ranges we tested for an individual hyperparameter. 

The \model{vgg16\_keras} network achieved 93.59\% accuracy on the original CIFAR-10 test set and 85.25\% accuracy on the new test set, a total gap of 8.34\%.
Through hyperparameter tuning, we found that the best configuration achieved an accuracy on the new test set of 85.84\%, a small improvement of 0.59\%.
Thus, tuning could account for only a small proportion of the accuracy gap.
}{}

\paragraph{Inspecting Hard Images}

Figure \ref{fig:hardtest} shows examples of the hard images in our new test set that no model correctly classified. 

\begin{figure}
\centering
\setlength{\imagedim}{2cm}
\setlength{\imagexspacing}{1cm}
\setlength{\imageyspacing}{1.5cm}
\newlength{\labelspacingtwo}
\setlength{\labelspacingtwo}{.2cm}
\centering
\begin{tikzpicture}
\iftoggle{isnips}{}{\footnotesize}
\tikzstyle{img}=[inner sep=0pt,outer sep=0pt];
\tikzstyle{imglabel}=[anchor=north,inner sep=0pt,yshift=-\labelspacingtwo];
\node [img] (image0) {\includegraphics[width=\imagedim]{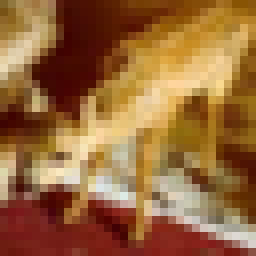}};
\node [img,anchor=west,at=(image0.east),xshift=\imagexspacing] (image1)
{\includegraphics[width=\imagedim]{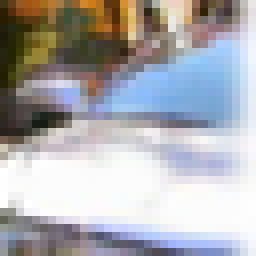}};
\node [img,anchor=west,at=(image1.east),xshift=\imagexspacing] (image2)
{\includegraphics[width=\imagedim]{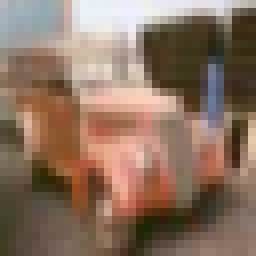}};
\node [img,anchor=west,at=(image2.east),xshift=\imagexspacing] (image3)
{\includegraphics[width=\imagedim]{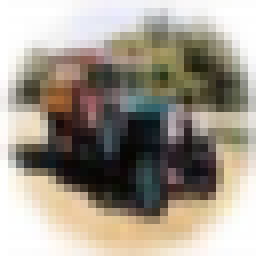}};
\node [img,anchor=west,at=(image3.east),xshift=\imagexspacing] (image4)
{\includegraphics[width=\imagedim]{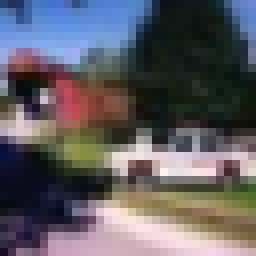}};
\node [img,anchor=north,at=(image0.south),yshift=-\imageyspacing] (image5)
{\includegraphics[width=\imagedim]{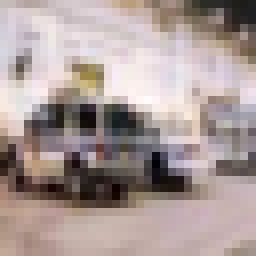}};
\node [img,anchor=west,at=(image5.east),xshift=\imagexspacing] (image6)
{\includegraphics[width=\imagedim]{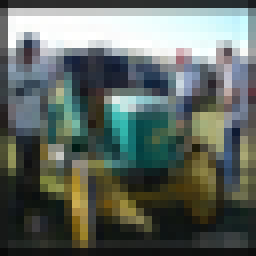}};
\node [img,anchor=west,at=(image6.east),xshift=\imagexspacing] (image7)
{\includegraphics[width=\imagedim]{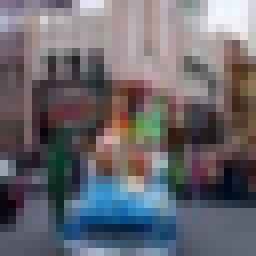}};
\node [img,anchor=west,at=(image7.east),xshift=\imagexspacing] (image8)
{\includegraphics[width=\imagedim]{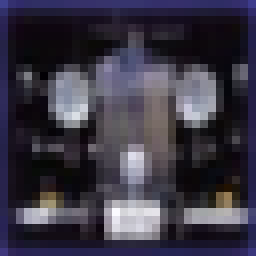}};
\node [img,anchor=west,at=(image8.east),xshift=\imagexspacing] (image9)
{\includegraphics[width=\imagedim]{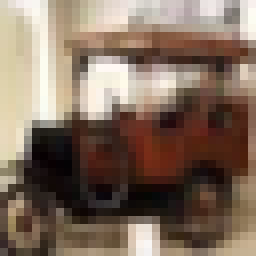}};
\node [img,anchor=north,at=(image5.south),yshift=-\imageyspacing] (image10)
{\includegraphics[width=\imagedim]{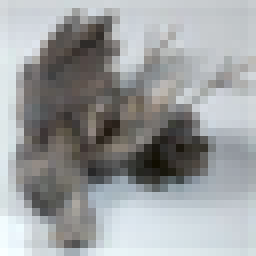}};
\node [img,anchor=west,at=(image10.east),xshift=\imagexspacing] (image11)
{\includegraphics[width=\imagedim]{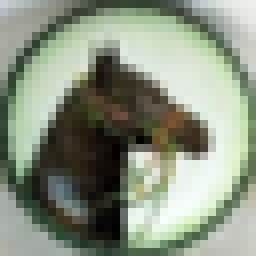}};
\node [img,anchor=west,at=(image11.east),xshift=\imagexspacing] (image12)
{\includegraphics[width=\imagedim]{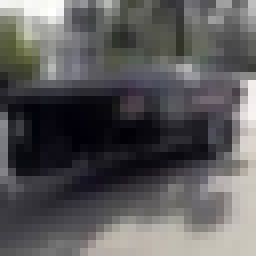}};
\node [img,anchor=west,at=(image12.east),xshift=\imagexspacing] (image13)
{\includegraphics[width=\imagedim]{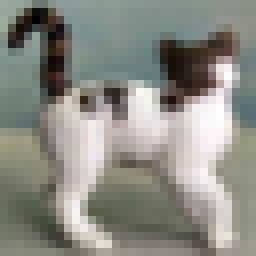}};
\node [img,anchor=west,at=(image13.east),xshift=\imagexspacing] (image14)
{\includegraphics[width=\imagedim]{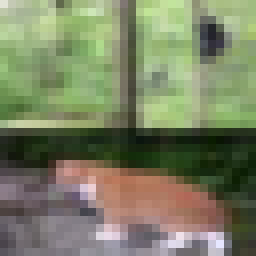}};
\node [img,anchor=north,at=(image10.south),yshift=-\imageyspacing] (image15)
{\includegraphics[width=\imagedim]{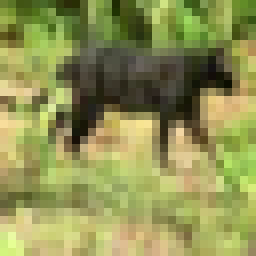}};
\node [img,anchor=west,at=(image15.east),xshift=\imagexspacing] (image16)
{\includegraphics[width=\imagedim]{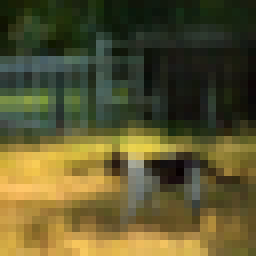}};
\node [img,anchor=west,at=(image16.east),xshift=\imagexspacing] (image17)
{\includegraphics[width=\imagedim]{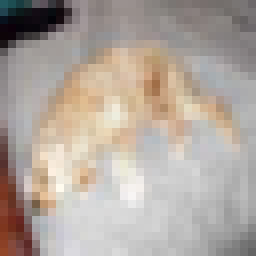}};
\node [img,anchor=west,at=(image17.east),xshift=\imagexspacing] (image18)
{\includegraphics[width=\imagedim]{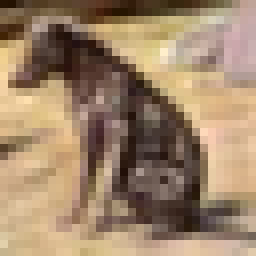}};
\node [img,anchor=west,at=(image18.east),xshift=\imagexspacing] (image19)
{\includegraphics[width=\imagedim]{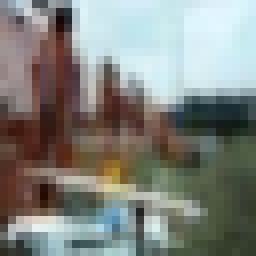}};
\node [img,anchor=north,at=(image15.south),yshift=-\imageyspacing] (image20)
{\includegraphics[width=\imagedim]{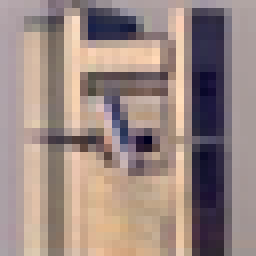}};
\node [imglabel] (label0) at (image0.south) [align=left]{True: \class{deer}\\ Predicted: \class{frog}};
\node [imglabel] (label1) at (image1.south) [align=left]{True: \class{automobile}\\ Predicted:
\class{airplane}};
\node [imglabel] (label2) at (image2.south) [align=left]{True: \class{automobile}\\ Predicted:
\class{truck}};
\node [imglabel] (label3) at (image3.south) [align=left]{True: \class{automobile}\\ Predicted: \class{truck}};
\node [imglabel] (label4) at (image4.south) [align=left]{True: \class{automobile}\\ Predicted: \class{airplane}};
\node [imglabel] (label5) at (image5.south) [align=left]{True: \class{automobile}\\ Predicted: \class{truck}};
\node [imglabel] (label6) at (image6.south) [align=left]{True: \class{automobile}\\ Predicted: \class{truck}};
\node [imglabel] (label7) at (image7.south) [align=left]{True: \class{automobile}\\ Predicted: \class{truck}};
\node [imglabel] (label8) at (image8.south) [align=left]{True: \class{automobile}\\ Predicted: \class{truck}};
\node [imglabel] (label9) at (image9.south) [align=left]{True: \class{automobile}\\ Predicted: \class{truck}};
\node [imglabel] (label10) at (image10.south) [align=left]{True: \class{bird}\\ Predicted: \class{frog}};
\node [imglabel] (label11) at (image11.south) [align=left]{True: \class{horse}\\ Predicted: \class{frog}};
\node [imglabel] (label12) at (image12.south) [align=left]{True: \class{ship}\\ Predicted: \class{automobile}};
\node [imglabel] (label13) at (image13.south) [align=left]{True: \class{cat}\\ Predicted: \class{dog}};
\node [imglabel] (label14) at (image14.south) [align=left]{True: \class{cat}\\ Predicted: \class{deer}};
\node [imglabel] (label15) at (image15.south) [align=left]{True: \class{cat}\\ Predicted: \class{deer}};
\node [imglabel] (label16) at (image16.south) [align=left]{True: \class{cat}\\ Predicted: \class{deer}};
\node [imglabel] (label17) at (image17.south) [align=left]{True: \class{dog}\\ Predicted: \class{cat}};
\node [imglabel] (label18) at (image18.south) [align=left]{True: \class{dog}\\ Predicted: \class{cat}};
\node [imglabel] (label19) at (image19.south) [align=left]{True: \class{airplane}\\ Predicted: \class{ship}};
\node [imglabel] (label20) at (image20.south) [align=left]{True: \class{airplane}\\ Predicted: \class{cat}};
\end{tikzpicture}
  \caption{Hard images that no model correctly classified in the new test set.  
  True gives the correct class label, while Predicted gives the label predicted by the majority of models. }
  \label{fig:hardtest}
\end{figure}

\iftoggle{isnips}{
\paragraph{Training on part of our new test set.}
We randomly drew a class-balanced split containing 1010 images from the new test set.  We then added these images to the full CIFAR-10 training set and retrained the \model{vgg16\_keras} model.  After training, we tested the model on the 1011 held-out images from the new test set.  We repeated this experiment twice with different randomly selected splits from our test set, obtaining accuracies of 85.06\% and 85.36\%. 
These numbers are all within the confidence interval of the accuracy reported for \model{vgg16\_keras} in Table \ref{tab:v4_results}. 

\paragraph{Cross Validation. }
We created cross-validation data by randomly dividing the training set into 5 class-balanced splits.  
A new training set was created by randomly shuffling together 4 out of 5 training splits along with the original test split.  
A new test set was created from the leftover held-out split from the training set.  
We then retrained the models \model{vgg\_15\_BN\_64}, \model{wide\_resnet\_28\_10}, and \model{shake\_shake\_64d\_cutout} on each of the 5 new datasets we created.  
The accuracies are reported in Table \ref{tab:cross_validation}.  The accuracies on each of the cross validation splits did not vary much from the accuracies on the original test set.  

\begin{table}[h!]
\centering
\begin{tabular}{c | c | c | c}
 & \model{vgg\_15\_BN\_64} & \model{wide\_resnet\_28\_10} & \model{shake\_shake\_64d\_cutout} \\ \hline
Split 1 & 93.87 & 96.16 &	97.16 \\
Split 2 & 93.81	& 96.04 &	97.3\\
Split 3 & 94.01	& 96.37 &	97.38\\
Split 4 & 93.99	& 96.16 &	97.39\\
Split 5 & 93.5	& 96.5  & 	97.4\\
\end{tabular}
\caption{Model Accuracies on cross validation splits}
\label{tab:cross_validation}
\end{table}
}
\pagebreak

\section{Error ratios}
\label{apx:v4_error_results}
\begin{table}[h!]
  \centering
  \rowcolors{2}{white}{gray!25}
  \begin{tabular}{lllr}
\toprule
{} &                                       Original Error &                                            New Error & Error Ratio \\
\midrule
\model{shake\_shake\_64d\_cutout}   \cite{shakeshake, cutout} &  2.9 {\footnotesize \textcolor{gray}{[3.2, 2.6]}} &  7.0 {\footnotesize \textcolor{gray}{[8.2, 6.0]}} &  2.4 \\
\model{shake\_shake\_96d}   \cite{shakeshake}                 &  2.9 {\footnotesize \textcolor{gray}{[3.3, 2.6]}} &  8.1 {\footnotesize \textcolor{gray}{[9.3, 6.9]}} &  2.8 \\
\model{shake\_shake\_64d}   \cite{shakeshake}                 &  3.0 {\footnotesize \textcolor{gray}{[3.4, 2.7]}} &  8.6 {\footnotesize \textcolor{gray}{[9.9, 7.4]}} &  2.8 \\
\model{wide\_resnet\_28\_10\_cutout}   \cite{wrn, cutout}     &  3.0 {\footnotesize \textcolor{gray}{[3.4, 2.7]}} &  8.0 {\footnotesize \textcolor{gray}{[9.3, 6.9]}} &  2.6 \\
\model{shake\_drop}   \cite{shakedrop}                        &  3.1 {\footnotesize \textcolor{gray}{[3.5, 2.8]}} &  7.7 {\footnotesize \textcolor{gray}{[9.0, 6.6]}} &  2.5 \\
\model{shake\_shake\_32d}   \cite{shakeshake}                 &  3.4 {\footnotesize \textcolor{gray}{[3.8, 3.1]}} &  10.2 {\footnotesize \textcolor{gray}{[11.6, 8.9]}} &  3 \\
\model{darc}   \cite{darc}                                    &  3.4 {\footnotesize \textcolor{gray}{[3.8, 3.1]}} &  10.5 {\footnotesize \textcolor{gray}{[11.9, 9.2]}} &  3.1 \\
\model{resnext\_29\_4x64d}   \cite{resnext}                   &  3.6 {\footnotesize \textcolor{gray}{[4.0, 3.3]}} &  10.4 {\footnotesize \textcolor{gray}{[11.8, 9.1]}} &  2.9 \\
\model{pyramidnet\_basic\_110\_270}   \cite{pyramidnet}       &  3.7 {\footnotesize \textcolor{gray}{[4.0, 3.3]}} &  9.5 {\footnotesize \textcolor{gray}{[10.9, 8.3]}} &  2.6 \\
\model{resnext\_29\_8x64d}   \cite{resnext}                   &  3.8 {\footnotesize \textcolor{gray}{[4.2, 3.4]}} &  10.0 {\footnotesize \textcolor{gray}{[11.4, 8.8]}} &  2.7 \\
\model{wide\_resnet\_28\_10}   \cite{wrn}                     &  4.1 {\footnotesize \textcolor{gray}{[4.5, 3.7]}} &  10.3 {\footnotesize \textcolor{gray}{[11.7, 9.0]}} &  2.5 \\
\model{pyramidnet\_basic\_110\_84}   \cite{pyramidnet}        &  4.3 {\footnotesize \textcolor{gray}{[4.7, 3.9]}} &  10.7 {\footnotesize \textcolor{gray}{[12.2, 9.4]}} &  2.5 \\
\model{densenet\_BC\_100\_12}   \cite{densenet}               &  4.5 {\footnotesize \textcolor{gray}{[4.9, 4.1]}} &  12.4 {\footnotesize \textcolor{gray}{[13.9, 11.0]}} &  2.8 \\
\model{neural\_architecture\_search}   \cite{nas}             &  4.6 {\footnotesize \textcolor{gray}{[5.0, 4.2]}} &  11.2 {\footnotesize \textcolor{gray}{[12.6, 9.8]}} &  2.4 \\
\model{wide\_resnet\_tf}   \cite{wrn}                         &  5.0 {\footnotesize \textcolor{gray}{[5.4, 4.6]}} &  11.5 {\footnotesize \textcolor{gray}{[13.0, 10.1]}} &  2.3 \\
\model{resnet\_v2\_bottleneck\_164}   \cite{resnet_preact}    &  5.8 {\footnotesize \textcolor{gray}{[6.3, 5.4]}} &  14.1 {\footnotesize \textcolor{gray}{[15.7, 12.6]}} &  2.4 \\
\model{vgg16\_keras}   \cite{vgg, vgg_cifar}                  &  6.4 {\footnotesize \textcolor{gray}{[6.9, 5.9]}} &  14.7 {\footnotesize \textcolor{gray}{[16.4, 13.2]}} &  2.3 \\
\model{resnet\_basic\_110}   \cite{resnet}                    &  6.5 {\footnotesize \textcolor{gray}{[7.0, 6.1]}} &  14.8 {\footnotesize \textcolor{gray}{[16.5, 13.3]}} &  2.3 \\
\model{resnet\_v2\_basic\_110}   \cite{resnet_preact}         &  6.6 {\footnotesize \textcolor{gray}{[7.1, 6.1]}} &  13.5 {\footnotesize \textcolor{gray}{[15.1, 12.0]}} &  2 \\
\model{resnet\_basic\_56}   \cite{resnet}                     &  6.7 {\footnotesize \textcolor{gray}{[7.2, 6.2]}} &  15.0 {\footnotesize \textcolor{gray}{[16.7, 13.5]}} &  2.2 \\
\model{resnet\_basic\_44}   \cite{resnet}                     &  7.0 {\footnotesize \textcolor{gray}{[7.5, 6.5]}} &  15.8 {\footnotesize \textcolor{gray}{[17.4, 14.2]}} &  2.3 \\
\model{vgg\_15\_BN\_64}   \cite{vgg, vgg_cifar}               &  7.0 {\footnotesize \textcolor{gray}{[7.5, 6.5]}} &  15.1 {\footnotesize \textcolor{gray}{[16.8, 13.6]}} &  2.2 \\
\model{resnet\_preact\_tf}   \cite{resnet}                    &  7.3 {\footnotesize \textcolor{gray}{[7.8, 6.8]}} &  15.6 {\footnotesize \textcolor{gray}{[17.3, 14.1]}} &  2.1 \\
\model{resnet\_basic\_32}   \cite{resnet}                     &  7.5 {\footnotesize \textcolor{gray}{[8.0, 7.0]}} &  15.1 {\footnotesize \textcolor{gray}{[16.8, 13.6]}} &  2 \\
\model{cudaconvnet}   \cite{alexnet}                          &  11.5 {\footnotesize \textcolor{gray}{[12.1, 10.8]}} &  22.5 {\footnotesize \textcolor{gray}{[24.3, 20.7]}} &  2 \\
\model{random\_features\_256k\_aug}   \cite{rf}               &  14.4 {\footnotesize \textcolor{gray}{[15.1, 13.7]}} &  26.9 {\footnotesize \textcolor{gray}{[28.9, 24.9]}} &  1.9 \\
\model{random\_features\_32k\_aug}   \cite{rf}                &  15.0 {\footnotesize \textcolor{gray}{[15.7, 14.3]}} &  28.1 {\footnotesize \textcolor{gray}{[30.1, 26.1]}} &  1.9 \\
\model{random\_features\_256k}   \cite{rf}                    &  15.8 {\footnotesize \textcolor{gray}{[16.5, 15.1]}} &  30.1 {\footnotesize \textcolor{gray}{[32.2, 28.1]}} &  1.9 \\
\model{random\_features\_32k}   \cite{rf}                     &  16.7 {\footnotesize \textcolor{gray}{[17.4, 16.0]}} &  32.1 {\footnotesize \textcolor{gray}{[34.1, 30.0]}} &  1.9 \\
\model{alexnet\_tf}                                           &  18.0 {\footnotesize \textcolor{gray}{[18.8, 17.3]}} &  31.1 {\footnotesize \textcolor{gray}{[33.2, 29.1]}} &  1.7 \\
\bottomrule
\end{tabular}

  \caption{Model error on the original CIFAR-10 test set and the new test set, as well as the
error ratio between the new and original test set. }
  \label{tab:v4_error_results}
\end{table} 
\pagebreak

\section{Class-balanced Test Set}
The top 25 keywords in the CIFAR-10 dataset capture approximately 95\% of the new dataset.  
However, the 5\% of the CIFAR-10 dataset that is not captured with the top 25 keywords is heavily skewed to the ship class. 
As a result, our new dataset was not precisely class-balanced, and contained only 8\% ships.
 
We created a class-balanced version of the new test set with exactly 2000 images. 
In this version, we selected the top 50 keywords in each class, and then computed a fractional number of images for that keyword based on the fractional ratio of that keyword in the corresponding CIFAR-10 class multiplied by the desired number of images for each class.  
We then sorted the fractional keyword counts by their rounding error, and removed one image from the keywords with the largest rounding gap until we had 2000 images.
The resulting dataset achieves slightly higher accuracy on the models, with the difference in accuracy from the new test set described in the main paper given in Table \ref{tab:v6_results}.   

\label{apx:v6_results}
\begin{table}[h!]
  \centering
  \rowcolors{2}{white}{gray!25}
  \begin{tabular}{lllll}
\toprule
{} &                                    Original Accuracy &                                         New Accuracy &  Gap & $\Delta$ Acc. \\
\midrule
\model{shake\_shake\_64d\_cutout}   \cite{shakeshake, cutout} &  97.1 {\footnotesize \textcolor{gray}{[96.8, 97.4]}} &  93.1 {\footnotesize \textcolor{gray}{[90.9, 93.3]}} &  4 &  0.13 \\
\model{shake\_shake\_96d}   \cite{shakeshake}                 &  97.1 {\footnotesize \textcolor{gray}{[96.7, 97.4]}} &  92.0 {\footnotesize \textcolor{gray}{[89.7, 92.2]}} &  5.1 &  0.015 \\
\model{shake\_shake\_64d}   \cite{shakeshake}                 &  97.0 {\footnotesize \textcolor{gray}{[96.6, 97.3]}} &  91.9 {\footnotesize \textcolor{gray}{[89.6, 92.2]}} &  5.1 &  0.46 \\
\model{wide\_resnet\_28\_10\_cutout}   \cite{wrn, cutout}     &  97.0 {\footnotesize \textcolor{gray}{[96.6, 97.3]}} &  92.1 {\footnotesize \textcolor{gray}{[89.8, 92.3]}} &  4.9 &  0.12 \\
\model{shake\_drop}   \cite{shakedrop}                        &  96.9 {\footnotesize \textcolor{gray}{[96.5, 97.2]}} &  92.3 {\footnotesize \textcolor{gray}{[90.0, 92.5]}} &  4.6 &  0.019 \\
\model{shake\_shake\_32d}   \cite{shakeshake}                 &  96.6 {\footnotesize \textcolor{gray}{[96.2, 96.9]}} &  90.0 {\footnotesize \textcolor{gray}{[87.6, 90.4]}} &  6.6 &  0.19 \\
\model{darc}   \cite{darc}                                    &  96.6 {\footnotesize \textcolor{gray}{[96.2, 96.9]}} &  89.9 {\footnotesize \textcolor{gray}{[87.5, 90.3]}} &  6.7 &  0.39 \\
\model{resnext\_29\_4x64d}   \cite{resnext}                   &  96.4 {\footnotesize \textcolor{gray}{[96.0, 96.7]}} &  90.1 {\footnotesize \textcolor{gray}{[87.8, 90.5]}} &  6.2 &  0.54 \\
\model{pyramidnet\_basic\_110\_270}   \cite{pyramidnet}       &  96.3 {\footnotesize \textcolor{gray}{[96.0, 96.7]}} &  90.5 {\footnotesize \textcolor{gray}{[88.1, 90.9]}} &  5.8 &  0.05 \\
\model{resnext\_29\_8x64d}   \cite{resnext}                   &  96.2 {\footnotesize \textcolor{gray}{[95.8, 96.6]}} &  90.1 {\footnotesize \textcolor{gray}{[87.7, 90.5]}} &  6.1 &  0.14 \\
\model{wide\_resnet\_28\_10}   \cite{wrn}                     &  95.9 {\footnotesize \textcolor{gray}{[95.5, 96.3]}} &  90.1 {\footnotesize \textcolor{gray}{[87.8, 90.5]}} &  5.8 &  0.44 \\
\model{pyramidnet\_basic\_110\_84}   \cite{pyramidnet}        &  95.7 {\footnotesize \textcolor{gray}{[95.3, 96.1]}} &  89.6 {\footnotesize \textcolor{gray}{[87.2, 90.0]}} &  6.1 &  0.34 \\
\model{densenet\_BC\_100\_12}   \cite{densenet}               &  95.5 {\footnotesize \textcolor{gray}{[95.1, 95.9]}} &  87.9 {\footnotesize \textcolor{gray}{[85.4, 88.4]}} &  7.6 &  0.32 \\
\model{neural\_architecture\_search}   \cite{nas}             &  95.4 {\footnotesize \textcolor{gray}{[95.0, 95.8]}} &  89.2 {\footnotesize \textcolor{gray}{[86.8, 89.6]}} &  6.2 &  0.38 \\
\model{wide\_resnet\_tf}   \cite{wrn}                         &  95.0 {\footnotesize \textcolor{gray}{[94.6, 95.4]}} &  88.8 {\footnotesize \textcolor{gray}{[86.4, 89.3]}} &  6.2 &  0.33 \\
\model{resnet\_v2\_bottleneck\_164}   \cite{resnet_preact}    &  94.2 {\footnotesize \textcolor{gray}{[93.7, 94.6]}} &  86.1 {\footnotesize \textcolor{gray}{[83.6, 86.7]}} &  8.1 &  0.2 \\
\model{vgg16\_keras}   \cite{vgg, vgg_cifar}                  &  93.6 {\footnotesize \textcolor{gray}{[93.1, 94.1]}} &  85.6 {\footnotesize \textcolor{gray}{[83.1, 86.3]}} &  8 &  0.35 \\
\model{resnet\_basic\_110}   \cite{resnet}                    &  93.5 {\footnotesize \textcolor{gray}{[93.0, 93.9]}} &  85.4 {\footnotesize \textcolor{gray}{[82.9, 86.1]}} &  8.1 &  0.24 \\
\model{resnet\_v2\_basic\_110}   \cite{resnet_preact}         &  93.4 {\footnotesize \textcolor{gray}{[92.9, 93.9]}} &  86.9 {\footnotesize \textcolor{gray}{[84.4, 87.5]}} &  6.5 &  0.41 \\
\model{resnet\_basic\_56}   \cite{resnet}                     &  93.3 {\footnotesize \textcolor{gray}{[92.8, 93.8]}} &  84.9 {\footnotesize \textcolor{gray}{[82.3, 85.5]}} &  8.5 & -0.11 \\
\model{resnet\_basic\_44}   \cite{resnet}                     &  93.0 {\footnotesize \textcolor{gray}{[92.5, 93.5]}} &  84.8 {\footnotesize \textcolor{gray}{[82.2, 85.5]}} &  8.2 &  0.58 \\
\model{vgg\_15\_BN\_64}   \cite{vgg, vgg_cifar}               &  93.0 {\footnotesize \textcolor{gray}{[92.5, 93.5]}} &  85.0 {\footnotesize \textcolor{gray}{[82.5, 85.7]}} &  7.9 &  0.19 \\
\model{resnet\_preact\_tf}   \cite{resnet}                    &  92.7 {\footnotesize \textcolor{gray}{[92.2, 93.2]}} &  85.1 {\footnotesize \textcolor{gray}{[82.6, 85.8]}} &  7.6 &  0.74 \\
\model{resnet\_basic\_32}   \cite{resnet}                     &  92.5 {\footnotesize \textcolor{gray}{[92.0, 93.0]}} &  85.2 {\footnotesize \textcolor{gray}{[82.7, 85.9]}} &  7.3 &  0.34 \\
\model{cudaconvnet}   \cite{alexnet}                          &  88.5 {\footnotesize \textcolor{gray}{[87.9, 89.2]}} &  78.2 {\footnotesize \textcolor{gray}{[75.5, 79.2]}} &  10 &  0.66 \\
\model{random\_features\_256k\_aug}   \cite{rf}               &  85.6 {\footnotesize \textcolor{gray}{[84.9, 86.3]}} &  73.6 {\footnotesize \textcolor{gray}{[70.8, 74.8]}} &  12 &  0.47 \\
\model{random\_features\_32k\_aug}   \cite{rf}                &  85.0 {\footnotesize \textcolor{gray}{[84.3, 85.7]}} &  72.2 {\footnotesize \textcolor{gray}{[69.4, 73.4]}} &  13 &  0.26 \\
\model{random\_features\_256k}   \cite{rf}                    &  84.2 {\footnotesize \textcolor{gray}{[83.5, 84.9]}} &  70.5 {\footnotesize \textcolor{gray}{[67.7, 71.7]}} &  14 &  0.58 \\
\model{random\_features\_32k}   \cite{rf}                     &  83.3 {\footnotesize \textcolor{gray}{[82.6, 84.0]}} &  68.7 {\footnotesize \textcolor{gray}{[65.9, 70.0]}} &  15 &  0.76 \\
\model{alexnet\_tf}                                           &  82.0 {\footnotesize \textcolor{gray}{[81.2, 82.7]}} &  69.2 {\footnotesize \textcolor{gray}{[66.4, 70.5]}} &  13 &  0.32 \\
\bottomrule
\end{tabular}

  \caption{Model accuracy on the original CIFAR-10 test set and the class-balanced new test set. Gap is
the difference between the original test set and the class-balanced test set. $\Delta$ Acc. is the
accuracy on the class balanced test set minus the new test set described in main paper.
}
  \label{tab:v6_results}
\end{table}
\pagebreak

\section{Keywords}
\label{apx:v4_keywords}
\captionsetup[subtable]{labelformat=empty}
\captionsetup[subtable]{position=top}
\begin{table}[h!]
  \captionsetup[table]{position=top}
  \caption{Distribution of the top 25 keywords in each class for the new and original test set.}
  \begin{subtable}{0.3\linewidth}
     \centering
     \iftoggle{isnips}{}{\scriptsize}
     \begin{tabular}{L{4cm}R{1cm}R{1cm}}
\toprule
\multicolumn{3}{c}{\textbf{Frog}} \\
\midrule
{} &   New &  Original \\
\midrule
bufo\_bufo           & 0.64\% &     0.63\% \\
leopard\_frog        & 0.64\% &     0.64\% \\
bufo\_viridis        & 0.59\% &     0.57\% \\
rana\_temporaria     & 0.54\% &     0.53\% \\
bufo                & 0.49\% &     0.47\% \\
bufo\_americanus     & 0.49\% &     0.46\% \\
toad                & 0.49\% &     0.46\% \\
green\_frog          & 0.45\% &     0.44\% \\
rana\_catesbeiana    & 0.45\% &     0.43\% \\
bufo\_marinus        & 0.45\% &     0.43\% \\
bullfrog            & 0.45\% &     0.42\% \\
american\_toad       & 0.45\% &     0.43\% \\
frog                & 0.35\% &     0.35\% \\
rana\_pipiens        & 0.35\% &     0.32\% \\
toad\_frog           & 0.30\% &     0.30\% \\
spadefoot           & 0.30\% &     0.27\% \\
western\_toad        & 0.30\% &     0.26\% \\
grass\_frog          & 0.30\% &     0.27\% \\
pickerel\_frog       & 0.25\% &     0.24\% \\
spring\_frog         & 0.25\% &     0.22\% \\
rana\_clamitans      & 0.20\% &     0.20\% \\
natterjack          & 0.20\% &     0.17\% \\
crapaud             & 0.20\% &     0.18\% \\
bufo\_calamita       & 0.20\% &     0.18\% \\
alytes\_obstetricans & 0.20\% &     0.16\% \\
\bottomrule
\end{tabular}

     \label{tab:frog}
  \end{subtable}%
  \hspace*{10em}
  \begin{subtable}{0.3\linewidth}
     \centering
     \iftoggle{isnips}{}{\scriptsize}
     \begin{tabular}{L{4cm}R{1cm}R{1cm}}
\toprule
\multicolumn{3}{c}{\textbf{Cat}} \\
\midrule
{} &   New &  Original \\
\midrule
tabby\_cat         & 1.78\% &     1.78\% \\
tabby             & 1.53\% &     1.52\% \\
domestic\_cat      & 1.34\% &     1.33\% \\
cat               & 1.24\% &     1.25\% \\
house\_cat         & 0.79\% &     0.79\% \\
felis\_catus       & 0.69\% &     0.69\% \\
mouser            & 0.64\% &     0.63\% \\
felis\_domesticus  & 0.54\% &     0.50\% \\
true\_cat          & 0.49\% &     0.47\% \\
tomcat            & 0.49\% &     0.49\% \\
alley\_cat         & 0.30\% &     0.30\% \\
felis\_bengalensis & 0.15\% &     0.11\% \\
nougat            & 0.10\% &     0.05\% \\
gray              & 0.05\% &     0.03\% \\
manx\_cat          & 0.05\% &     0.04\% \\
fissiped          & 0.05\% &     0.03\% \\
persian\_cat       & 0.05\% &     0.03\% \\
puss              & 0.05\% &     0.05\% \\
catnap            & 0.05\% &     0.03\% \\
tiger\_cat         & 0.05\% &     0.03\% \\
black\_cat         & 0.05\% &     0.04\% \\
bedspread         & 0.00\% &     0.02\% \\
siamese\_cat       & 0.00\% &     0.02\% \\
tortoiseshell     & 0.00\% &     0.02\% \\
kitty-cat         & 0.00\% &     0.02\% \\
\bottomrule
\end{tabular}

     \label{tab:cat} 
  \end{subtable} \\
  \begin{subtable}{0.3\linewidth}
     \centering
     \iftoggle{isnips}{}{\scriptsize}
     \begin{tabular}{L{4cm}R{1cm}R{1cm}}
\toprule
\multicolumn{3}{c}{\textbf{Dog}} \\
\midrule
\toprule
{} &   New &  Original \\
\midrule
pekingese            & 1.24\% &     1.22\% \\
maltese              & 0.94\% &     0.93\% \\
puppy                & 0.89\% &     0.87\% \\
chihuahua            & 0.84\% &     0.81\% \\
dog                  & 0.69\% &     0.67\% \\
pekinese             & 0.69\% &     0.66\% \\
toy\_spaniel          & 0.59\% &     0.60\% \\
mutt                 & 0.49\% &     0.47\% \\
mongrel              & 0.49\% &     0.49\% \\
maltese\_dog          & 0.45\% &     0.43\% \\
toy\_dog              & 0.40\% &     0.36\% \\
japanese\_spaniel     & 0.40\% &     0.38\% \\
blenheim\_spaniel     & 0.35\% &     0.35\% \\
english\_toy\_spaniel  & 0.35\% &     0.31\% \\
domestic\_dog         & 0.35\% &     0.32\% \\
peke                 & 0.30\% &     0.28\% \\
canis\_familiaris     & 0.30\% &     0.27\% \\
lapdog               & 0.30\% &     0.30\% \\
king\_charles\_spaniel & 0.20\% &     0.17\% \\
toy                  & 0.15\% &     0.13\% \\
feist                & 0.10\% &     0.06\% \\
pet                  & 0.10\% &     0.07\% \\
cavalier             & 0.10\% &     0.05\% \\
canine               & 0.05\% &     0.04\% \\
cur                  & 0.05\% &     0.04\% \\
\bottomrule
\end{tabular}

     \label{tab:dog}
  \end{subtable}
  \hspace*{10em}
  \begin{subtable}{0.3\linewidth}
     \centering
     \iftoggle{isnips}{}{\scriptsize}
     \begin{tabular}{L{4cm}R{1cm}R{1cm}}
\toprule
\multicolumn{3}{c}{\textbf{Deer}} \\
\midrule
{} &   New &  Original \\
\midrule
elk                 & 0.79\% &     0.77\% \\
capreolus\_capreolus & 0.74\% &     0.71\% \\
cervus\_elaphus      & 0.64\% &     0.61\% \\
fallow\_deer         & 0.64\% &     0.63\% \\
roe\_deer            & 0.59\% &     0.60\% \\
deer                & 0.59\% &     0.60\% \\
muntjac             & 0.54\% &     0.51\% \\
mule\_deer           & 0.54\% &     0.51\% \\
odocoileus\_hemionus & 0.49\% &     0.50\% \\
fawn                & 0.49\% &     0.49\% \\
alces\_alces         & 0.40\% &     0.36\% \\
wapiti              & 0.40\% &     0.36\% \\
american\_elk        & 0.40\% &     0.35\% \\
red\_deer            & 0.35\% &     0.33\% \\
moose               & 0.35\% &     0.35\% \\
rangifer\_caribou    & 0.25\% &     0.24\% \\
rangifer\_tarandus   & 0.25\% &     0.24\% \\
caribou             & 0.25\% &     0.23\% \\
sika                & 0.25\% &     0.22\% \\
woodland\_caribou    & 0.25\% &     0.21\% \\
dama\_dama           & 0.20\% &     0.19\% \\
cervus\_sika         & 0.20\% &     0.16\% \\
barking\_deer        & 0.20\% &     0.18\% \\
sambar              & 0.15\% &     0.15\% \\
stag                & 0.15\% &     0.13\% \\
\bottomrule
\end{tabular}

     \label{tab:deer}
  \end{subtable} \\
\end{table} 
 \pagebreak
\begin{table*}  
  \begin{subtable}{0.3\linewidth}
     \centering
     \iftoggle{isnips}{}{\scriptsize}
     \begin{tabular}{L{4cm}R{1cm}R{1cm}}
\toprule
\multicolumn{3}{c}{\textbf{Bird}} \\
\midrule
{} &   New &  Original \\
\midrule
cassowary                & 0.89\% &     0.85\% \\
bird                     & 0.84\% &     0.84\% \\
wagtail                  & 0.74\% &     0.74\% \\
ostrich                  & 0.69\% &     0.68\% \\
struthio\_camelus         & 0.54\% &     0.51\% \\
sparrow                  & 0.54\% &     0.52\% \\
emu                      & 0.54\% &     0.51\% \\
pipit                    & 0.49\% &     0.47\% \\
passerine                & 0.49\% &     0.50\% \\
accentor                 & 0.49\% &     0.49\% \\
honey\_eater              & 0.40\% &     0.37\% \\
dunnock                  & 0.40\% &     0.37\% \\
alauda\_arvensis          & 0.30\% &     0.26\% \\
nandu                    & 0.30\% &     0.27\% \\
prunella\_modularis       & 0.30\% &     0.30\% \\
anthus\_pratensis         & 0.30\% &     0.28\% \\
finch                    & 0.25\% &     0.24\% \\
lark                     & 0.25\% &     0.20\% \\
meadow\_pipit             & 0.25\% &     0.20\% \\
rhea\_americana           & 0.25\% &     0.21\% \\
flightless\_bird          & 0.15\% &     0.10\% \\
emu\_novaehollandiae      & 0.15\% &     0.12\% \\
dromaius\_novaehollandiae & 0.15\% &     0.14\% \\
apteryx                  & 0.15\% &     0.10\% \\
flying\_bird              & 0.15\% &     0.13\% \\
\bottomrule
\end{tabular}

     \label{tab:bird}
  \end{subtable}
  \hspace*{10em}
  \begin{subtable}{0.3\linewidth}
     \centering
     \iftoggle{isnips}{}{\scriptsize}
     \begin{tabular}{L{4cm}R{1cm}R{1cm}}
\toprule
\multicolumn{3}{c}{\textbf{Ship}} \\
\midrule
{} &   New &  Original \\
\midrule
passenger\_ship   & 0.79\% &     0.78\% \\
boat             & 0.64\% &     0.64\% \\
cargo\_ship       & 0.40\% &     0.37\% \\
cargo\_vessel     & 0.40\% &     0.39\% \\
pontoon          & 0.35\% &     0.31\% \\
container\_ship   & 0.35\% &     0.31\% \\
speedboat        & 0.35\% &     0.32\% \\
freighter        & 0.35\% &     0.32\% \\
pilot\_boat       & 0.35\% &     0.31\% \\
ship             & 0.35\% &     0.31\% \\
cabin\_cruiser    & 0.30\% &     0.29\% \\
police\_boat      & 0.30\% &     0.25\% \\
sea\_boat         & 0.30\% &     0.29\% \\
oil\_tanker       & 0.30\% &     0.29\% \\
pleasure\_boat    & 0.25\% &     0.21\% \\
lightship        & 0.25\% &     0.22\% \\
powerboat        & 0.25\% &     0.25\% \\
guard\_boat       & 0.25\% &     0.20\% \\
dredger          & 0.25\% &     0.20\% \\
hospital\_ship    & 0.25\% &     0.21\% \\
banana\_boat      & 0.20\% &     0.19\% \\
merchant\_ship    & 0.20\% &     0.17\% \\
liberty\_ship     & 0.20\% &     0.15\% \\
container\_vessel & 0.20\% &     0.19\% \\
tanker           & 0.20\% &     0.18\% \\
\bottomrule
\end{tabular}

     \label{tab:ship}
  \end{subtable} \\
  \begin{subtable}{0.3\linewidth}
     \centering
     \iftoggle{isnips}{}{\scriptsize}
     \begin{tabular}{L{4cm}R{1cm}R{1cm}}
\toprule
\multicolumn{3}{c}{\textbf{Truck}} \\
\midrule
{} &   New &  Original \\
\midrule
dump\_truck          & 0.89\% &     0.89\% \\
trucking\_rig        & 0.79\% &     0.76\% \\
delivery\_truck      & 0.64\% &     0.61\% \\
truck               & 0.64\% &     0.65\% \\
tipper\_truck        & 0.64\% &     0.60\% \\
camion              & 0.59\% &     0.58\% \\
fire\_truck          & 0.59\% &     0.55\% \\
lorry               & 0.54\% &     0.53\% \\
garbage\_truck       & 0.54\% &     0.53\% \\
moving\_van          & 0.35\% &     0.32\% \\
tractor\_trailer     & 0.35\% &     0.34\% \\
tipper              & 0.35\% &     0.30\% \\
aerial\_ladder\_truck & 0.35\% &     0.34\% \\
ladder\_truck        & 0.30\% &     0.26\% \\
fire\_engine         & 0.30\% &     0.27\% \\
dumper              & 0.30\% &     0.28\% \\
trailer\_truck       & 0.30\% &     0.28\% \\
wrecker             & 0.30\% &     0.27\% \\
articulated\_lorry   & 0.25\% &     0.24\% \\
tipper\_lorry        & 0.25\% &     0.25\% \\
semi                & 0.20\% &     0.18\% \\
sound\_truck         & 0.15\% &     0.12\% \\
tow\_truck           & 0.15\% &     0.12\% \\
delivery\_van        & 0.15\% &     0.11\% \\
bookmobile          & 0.10\% &     0.10\% \\
\bottomrule
\end{tabular}

     \label{tab:truck}  
  \end{subtable}
  \hspace*{10em}
  \begin{subtable}{0.3\linewidth}
     \centering
     \iftoggle{isnips}{}{\scriptsize}
     \begin{tabular}{L{4cm}R{1cm}R{1cm}}
\toprule
\multicolumn{3}{c}{\textbf{Horse}} \\
\midrule
{} &   New &  Original \\
\midrule
arabian                 & 1.14\% &     1.12\% \\
lipizzan                & 1.04\% &     1.02\% \\
broodmare               & 0.99\% &     0.97\% \\
gelding                 & 0.74\% &     0.73\% \\
quarter\_horse           & 0.74\% &     0.72\% \\
stud\_mare               & 0.69\% &     0.69\% \\
lippizaner              & 0.54\% &     0.52\% \\
appaloosa               & 0.49\% &     0.45\% \\
lippizan                & 0.49\% &     0.46\% \\
dawn\_horse              & 0.45\% &     0.42\% \\
stallion                & 0.45\% &     0.43\% \\
tennessee\_walker        & 0.45\% &     0.45\% \\
tennessee\_walking\_horse & 0.40\% &     0.38\% \\
walking\_horse           & 0.30\% &     0.28\% \\
riding\_horse            & 0.20\% &     0.20\% \\
saddle\_horse            & 0.20\% &     0.18\% \\
female\_horse            & 0.15\% &     0.11\% \\
cow\_pony                & 0.15\% &     0.11\% \\
male\_horse              & 0.15\% &     0.14\% \\
buckskin                & 0.15\% &     0.13\% \\
horse                   & 0.10\% &     0.08\% \\
equine                  & 0.10\% &     0.08\% \\
quarter                 & 0.10\% &     0.07\% \\
cavalry\_horse           & 0.10\% &     0.09\% \\
thoroughbred            & 0.10\% &     0.06\% \\
\bottomrule
\end{tabular}

     \label{tab:horse}
  \end{subtable} \\
\end{table*}

\begin{table*}
  \begin{subtable}{0.3\linewidth}
     \centering
     \iftoggle{isnips}{}{\scriptsize}
     \begin{tabular}{L{4cm}R{1cm}R{1cm}}
\toprule
\multicolumn{3}{c}{\textbf{Airplane}} \\
\midrule
{} &   New &  Original \\
\midrule
stealth\_bomber       & 0.94\% &     0.92\% \\
airbus               & 0.89\% &     0.89\% \\
stealth\_fighter      & 0.79\% &     0.80\% \\
fighter\_aircraft     & 0.79\% &     0.76\% \\
biplane              & 0.74\% &     0.74\% \\
attack\_aircraft      & 0.69\% &     0.67\% \\
airliner             & 0.64\% &     0.61\% \\
jetliner             & 0.59\% &     0.56\% \\
monoplane            & 0.54\% &     0.55\% \\
twinjet              & 0.54\% &     0.52\% \\
dive\_bomber          & 0.54\% &     0.52\% \\
jumbo\_jet            & 0.49\% &     0.47\% \\
jumbojet             & 0.35\% &     0.35\% \\
propeller\_plane      & 0.30\% &     0.28\% \\
fighter              & 0.20\% &     0.20\% \\
plane                & 0.20\% &     0.15\% \\
amphibious\_aircraft  & 0.20\% &     0.20\% \\
multiengine\_airplane & 0.15\% &     0.14\% \\
seaplane             & 0.15\% &     0.14\% \\
floatplane           & 0.10\% &     0.05\% \\
multiengine\_plane    & 0.10\% &     0.06\% \\
reconnaissance\_plane & 0.10\% &     0.09\% \\
airplane             & 0.10\% &     0.08\% \\
tail                 & 0.10\% &     0.05\% \\
joint                & 0.05\% &     0.04\% \\
\bottomrule
\end{tabular}

     \label{tab:airplane}
  \end{subtable}
  \hspace*{10em}
  \begin{subtable}{0.3\linewidth}
     \centering
     \iftoggle{isnips}{}{\scriptsize}
     \begin{tabular}{L{4cm}R{1cm}R{1cm}}
\toprule
\multicolumn{3}{c}{\textbf{Automobile}} \\
\midrule
{} &   New &  Original \\
\midrule
coupe          & 1.29\% &     1.26\% \\
convertible    & 1.19\% &     1.18\% \\
station\_wagon  & 0.99\% &     0.98\% \\
automobile     & 0.89\% &     0.90\% \\
car            & 0.84\% &     0.81\% \\
auto           & 0.84\% &     0.83\% \\
compact\_car    & 0.79\% &     0.76\% \\
shooting\_brake & 0.64\% &     0.63\% \\
estate\_car     & 0.59\% &     0.59\% \\
wagon          & 0.54\% &     0.51\% \\
police\_cruiser & 0.45\% &     0.45\% \\
motorcar       & 0.40\% &     0.40\% \\
taxi           & 0.20\% &     0.17\% \\
cruiser        & 0.15\% &     0.13\% \\
compact        & 0.15\% &     0.11\% \\
beach\_wagon    & 0.15\% &     0.13\% \\
funny\_wagon    & 0.10\% &     0.05\% \\
gallery        & 0.10\% &     0.07\% \\
cab            & 0.10\% &     0.07\% \\
ambulance      & 0.10\% &     0.07\% \\
door           & 0.00\% &     0.03\% \\
ford           & 0.00\% &     0.03\% \\
opel           & 0.00\% &     0.03\% \\
sport\_car      & 0.00\% &     0.03\% \\
sports\_car     & 0.00\% &     0.03\% \\
\bottomrule
\end{tabular}

     \label{tab:Automobile}
  \end{subtable}
\end{table*}

\end{document}